\documentclass{article} 
\usepackage{collas2022_conference,times}

\usepackage{amsmath,amsfonts,bm}

\def\eqref#1{equation~\ref{#1}}

\def\1{\bm{1}}

\DeclareMathAlphabet{\mathsfit}{\encodingdefault}{\sfdefault}{m}{sl}
\SetMathAlphabet{\mathsfit}{bold}{\encodingdefault}{\sfdefault}{bx}{n}

\usepackage{sidecap}
\sidecaptionvpos{figure}{t}
\usepackage{url}
\usepackage{epsfig}
\usepackage{amsmath,amssymb}
\usepackage[utf8]{inputenc}
\usepackage{diagbox}
\usepackage{graphicx}
\usepackage{booktabs}
\usepackage{subfigure}
\usepackage{float}
\usepackage{xcolor}
\usepackage{algorithm}
\usepackage[noend]{algpseudocode}
\usepackage{multirow}
\usepackage[T1]{fontenc}
\usepackage{bbm}
\usepackage{colortbl}
\usepackage{enumerate}
\usepackage{soul}
\usepackage{pifont}
\newcommand{\cmark}{\ding{51}}%
\newcommand{\xmark}{\ding{55}}%

\usepackage{hyperref}
\hypersetup{
    colorlinks=true,
    linkcolor=red,
    filecolor=magenta,      
    urlcolor=blue,
    citecolor=purple,
    pdftitle={Overleaf Example},
    pdfpagemode=FullScreen,
    }

\makeatletter
\newcommand{\printfnsymbol}[1]{%
  \textsuperscript{\@fnsymbol{#1}}%
}

\title{SYNERgy between SYNaptic consolidation and Experience Replay for general continual learning}

\author{Fahad Sarfraz\thanks{Contributed equally.~~~~~~~~{$^1$ We open source our code at \url{https://github.com/NeurAI-Lab/SYNERgy}.}}, Elahe Arani\printfnsymbol{1}, Bahram Zonooz \\
Advanced Research Lab, NavInfo Europe, The Netherlands\\
\footnotesize\texttt{\{fahad.sarfraz,elahe.arani\}@navinfo.eu, bahram.zonooz@gmail.com}
}

\collasfinalcopy 

\begin{document}

\maketitle

\begin{abstract}
Continual learning (CL) in the brain is facilitated by a complex set of mechanisms. This includes the interplay of multiple memory systems for consolidating information as posited by the complementary learning systems (CLS) theory and synaptic consolidation for protecting the acquired knowledge from erasure. Thus, we propose a general CL method that creates a synergy between SYNaptic consolidation and dual memory Experience Replay (SYNERgy). Our method maintains a semantic memory that accumulates and consolidates information across the tasks and interacts with the episodic memory for effective replay. It further employs synaptic consolidation by tracking the importance of parameters during the training trajectory and anchoring them to the consolidated parameters in the semantic memory. To the best of our knowledge, our study is the first to employ dual memory experience replay in conjunction with synaptic consolidation that is suitable for general CL whereby the network does not utilize task boundaries or task labels during training or inference. Our evaluation on various challenging CL scenarios and characteristics analyses demonstrate the efficacy of incorporating both synaptic consolidation and CLS theory in enabling effective CL in DNNs.\footnotemark
\end{abstract}


\section{Introduction}

A major challenge towards achieving general intelligence and making DNNs suitable for deployment in an ever-changing environment is the ability to continuously acquire, retain and consolidate knowledge~\citep{parisi2019continual}. However, DNNs exhibit catastrophic forgetting whereby the network's performance on previously learned tasks drops drastically as they learn a new task~\citep{mccloskey1989catastrophic}. This can be majorly attributed to the violation of the strong i.i.d assumption in the sequential learning of tasks involved in CL~\citep{hadsell2020embracing}. Furthermore, catastrophic forgetting is also considered as an inevitable feature of the standard training procedure on connectionist models~\citep{kirkpatrick2017overcoming} whereby all the weights of the model are updated to learn a new task, resulting in overwriting of important weights for the previous tasks. To avoid catastrophic forgetting without hampering the learning of new tasks, a CL method needs to be plastic enough to adapt to the new data while being stable enough to not forget the already acquired knowledge.

In the brain, a delicate balance between plasticity and stability is maintained by a rich set of neurophysiological mechanisms \citep{parisi2019continual,zenke2017continual} and facilitated by multiple memory systems \citep{hassabis2017neuroscience}.
In particular, synaptic consolidation~\citep{clopath2012synaptic} retains information over longer periods by decreasing the rates of plasticity in a proportion of strengthened synapses that persist despite learning of other tasks \citep{cichon2015branch}. Furthermore, the CLS theory posits that intelligent agents must possess two complementary learning systems: instance-based hippocampal system and parametric neocortical system. The Hippocampus rapidly encodes novel information which is gradually consolidated into the structured knowledge representation of the neocortex through the replay of neural activity patterns that accompanied the learning event ~\citep{tulving2002episodic,kumaran2016learning}. The synergy between these mechanisms plays a crucial role in enabling efficient lifelong learning in the brain. Therefore, we hypothesize that an efficient method of employing a dual memory replay mechanism and incorporating synaptic consolidation for mediating the plasticity of weights can enable effective CL in DNNs.

Several works have taken inspiration from the studies in neuroscience, but they have mainly focused on a single aspect. The replay mechanism in Hippocampus has inspired a series of rehearsal-based approaches \citep{chaudhry2018efficient,lopez2017gradient} which maintains an episodic memory buffer and replays samples from previous tasks while training on the new task. On the other hand, regularization-based approaches \citep{kirkpatrick2017overcoming,ritter2018online,zenke2017continual} derive inspiration from synaptic consolidation and penalize changes in the network weights. Both of these approaches have their merits and demerits. Rehearsal-based approaches have proven to be more effective~\citep{farquhar2018towards} but require the amount of memory being stored and replayed to be proportional to the number of tasks which can be a limiting factor in learning longer sequences. On the other hand, while regularization-based approaches alleviate the need to store samples from previous tasks,  they fail to estimate the joint distribution of all the encountered tasks on challenging CL settings~\citep{farquhar2018towards} and often require information about the task boundaries and/or task labels. Research in both of these directions has mostly been orthogonal and an optimal method for incorporating both of these mechanisms for enabling CL in DNNs is not well studied. We, therefore, explore the design space of CL methods that employ experience replay to consolidate knowledge and synaptic consolidation to retain acquired knowledge over longer periods.

\begin{figure}[t]
    \centering
    \includegraphics[width=1\linewidth]{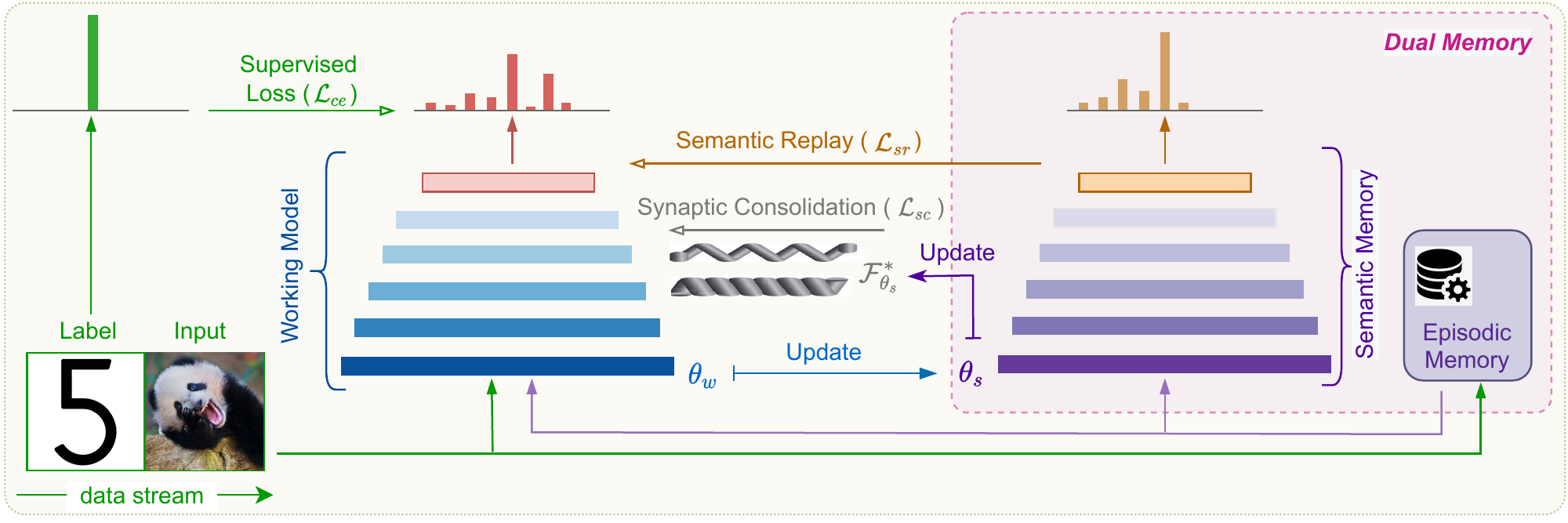}
    \caption{SYNERgy employs synaptic consolidation alongside an experience replay in a dual memory system. The semantic memory, $\theta_s$, is maintained by taking an exponential moving average of the working model, $\theta_w$, to aggregate knowledge across the tasks. The semantic memory then interacts with the episodic memory to extract consolidated replay logits for semantic replay in addition to episodic replay. This enables the working model to effectively consolidate knowledge across tasks by learning the joint distribution of tasks while aligning its decision boundary with the semantic memory. Furthermore, SYNERgy tracks the importance of the semantic memory parameters by aggregating Fisher information matrices, $\mathcal{F}_{\theta_S}^*$, and constrains the weight changes of the working model by anchoring them to the important parameters of the semantic memory through the synaptic consolidation loss $\mathcal{L}_{sc}$ (details in Algorithm \ref{algo:SYNERgy}).}
    \label{fig:synergy}
\end{figure}

We propose SYNERgy which employs a dual memory replay mechanism in conjunction with a synaptic consolidation for general incremental learning (General-IL) (Figure \ref{fig:synergy}). As the structural knowledge of the learned tasks is encoded in the weights of the working model~\citep{kirkpatrick2017overcoming}, our method maintains an additional semantic memory which aggregates and consolidates knowledge across the tasks by taking an exponential moving average of the working model's weights in a stochastic manner. The semantic memory interacts with the episodic memory to provide consolidated replay activations for enforcing a similarity structure on the update of the working model. This facilities knowledge consolidation and aligns the working model's decision boundary with the semantic memory. To further retain knowledge of previous tasks, SYNERgy incorporates synaptic consolidation by tracking the importance of parameters during the training trajectory and slowing down learning on parameters that are considered important for the acquired knowledge. Our work focuses on the realistic General-IL setting which involves training the method on a long sequence of tasks where the boundaries between the tasks are not distinct, the tasks themselves are not disjoint, and the method does not have access to task boundaries during training or testing~\citep{arani2022learning}. Therefore, SYNERgy does not utilize the task boundaries or make any assumption about the distribution of the data stream.

We extensively evaluate our method on a wide range of CL scenarios which covers a broad spectrum of the challenges a learning agent has to face in the real-world. The empirical results demonstrate the effectiveness and versatility of SYNERgy in acquiring, consolidating, and retaining knowledge over longer sequences. In addition to reducing forgetting, our method improves the reliability of the model, reduces the bias towards recent tasks, and converges to wider minima. Our work presents a compelling case for further exploring the design space of brain-inspired CL methods which incorporates both synaptic consolidation and complementary learning systems.

\section{Background}

\textbf{Regularization-based methods} draw inspiration from synaptic consolidation in the brain to propose algorithms which control the rate of learning on model weights based on their importance to previously seen tasks \citep{kirkpatrick2017overcoming}. oEWC~\citep{schwarz2018progress} estimates the importance of each parameter with the empirical Fisher information matrix (FIM), aggregates the FIMs evaluated at the task boundaries, and subsequently slows down learning on the subset of the model weights which are considered important for previous tasks. SI~\citep{zenke2017continual} computes an importance measure online along the entire learning trajectory. While it constitute a promising biologically plausible direction, they fail to learn the joint distribution of tasks and avoid catastrophic forgetting in more challenging settings~\citep{farquhar2018towards,hadsell2020embracing}. Additionally, they require task boundary and/or task label information which limits their application in the General-IL setting where task boundaries are not discrete.

\textbf{Rehearsal-based approaches} are inspired by the critical role that the replay of past neural activation patterns in the brain play in memory formation, consolidation, and retrieval. Experience Replay (ER)~~\citep{riemer2018learning} typically involves storing a subset of samples from previous tasks and mixing them with data from the new task to update the model parameters with an approximate joint distribution and has proven to be effective in reducing catastrophic forgetting under challenging settings~\citep{farquhar2018towards}. Several techniques have since been employed on top of ER to better utilize the memory samples \citep{lopez2017gradient,aljundi2019gradient}. 
Dark Experience Replay (DER++) samples logits during the entire optimization trajectory and adds a consistency loss on top of ER.
A key limitation of experience replay methods is that the memories being stored and replayed need to be proportional to the number of tasks as for a fixed buffer size, the representation of earlier tasks in memory diminishes as we learn over longer sequences. Furthermore, an optimal mechanism for replay is still an open question~\citep{hayes2021replay}.

\textbf{CLS theory} has further inspired dual memory learning systems \citep{pham2021dualnet,wang2022dualprompt}. 
CLS-ER \citep{arani2022learning} mimics the fast and slow learning in CLS by maintaining short- and long-term semantic memories which stochastically aggregates the weights of the working model using exponentially moving average at different rates. The semantic memories interact with the episodic memory to enforce a consistency loss for replay.
We use their proposed Mean-ER approach as it provides an effective and simpler approach for studying the effect of synaptic consolidation in a dual memory replay architecture.

While all these approaches have their merits and demerits and individually constitute a promising direction, research has been predominantly orthogonal. Our work draws inspiration from how these mechanisms together enable effective learning in the brain and explore the design space of CL methods that employ synaptic consolidation and experience replay in multiple memory systems for General-IL.

\section{Methodology}
The human brain has evolved to continually acquire, retain and consolidate knowledge. This ability to efficiently learn from an ever-changing environment is enabled by a rich set of neurophysiological mechanisms and multiple memory systems. Studies focusing on the structure and function of dendritic spines during learning show evidence for specialized mechanisms that protect knowledge about previous tasks from interference during learning on a new task~\citep{cichon2015branch,yang2009stably}. 
Neurobiological models suggest that CL in the neocortex relies on a process of task-specific synaptic consolidation which involves rendering a proportion of synapses less plastic and therefore stable over long timescales~\citep{benna2016computational,kirkpatrick2017overcoming}.
Another important characteristic of the brain is that it relies on multiple memory systems. The CLS theory ~\citep{kumaran2016learning} posits that intelligent agents must possess two learning systems: the Hippocampus quickly learns the specifics of individual experiences which are then gradually consolidated into the structured knowledge representations in the Neocortex. This process is accompanied by the replay of the structured patterns of neural activities that accompanied the learning event.
We believe that exploring the design space of models that employ both synaptic consolidation and CLS theory mechanisms may hold the key to enabling more effective CL in DNNs. To this end, similar to the brain, our proposed method SYNERgy incorporates synaptic consolidation and experience replay in a dual memory manner for General-IL. 


\subsection{Dual Memory System} 
Inspired by the interplay of the rapid instance-based system (Hippocampus) and the slow parametric system (Neocortex) in the brain, our method employs an instance-based episodic memory and a semantic memory which maintains consolidated structured knowledge across the learned tasks.

\textbf{Episodic Memory:}
To replay samples from the previous tasks, we utilize a small episodic memory buffer as an instance-based memory system which can be thought of as a very primitive hippocampus, permitting a form of complementary learning~\citep{hassabis2017neuroscience}. As our focus is on General-IL, we employ \textit{Reservoir sampling}~\citep{vitter1985random} to maintain the memory buffer which does not utilize the task boundaries to select the memory samples. It instead attempts to approximately match the distribution of the incoming stream by assigning equal probability to each incoming sample for being added to the buffer and replacing a random sample in the buffer. The probability for an incoming sample to be added to the buffer is given by $\frac{\mathcal{B}}{N_s}$, where $\mathcal{B}$ is the buffer size and $N_s$ is the total number of seen samples. Note that we opt for Reservoir sampling because of its simplicity, effectiveness, and suitability for General-IL, however, any method for memory selection and retrieval can be employed in our method. 

\textbf{Semantic Memory:}
The acquired knowledge of the tasks in DNNs is represented in the learned weights \citep{krishnan2019biologically}. Therefore, we aim to progressively accumulate and consolidate knowledge across the tasks encoded in the subsequent learned weights of the working model ($\theta_w$) throughout the training trajectory. To this end, we follow the approach in ~\citet{arani2022learning} to maintain our semantic memory (parametrized by $\theta_s$) by taking the exponential moving average of the working model as they provides an efficient approach for aggregating weights:
\begin{equation} \label{eq:sem_update}
    \theta_s \leftarrow \alpha_s\theta_s+(1-\alpha_s)~\theta_w,~~~~~if~~ r_s>a \sim U(0,1)
\end{equation}
where $\alpha_s$ is the decay parameter and $r_s$ is the update rate. As the semantic memory is designed to stochastically accumulate knowledge across the entire training trajectory by aggregating the weights of the working model, it can also be considered as an ensemble of several student models with varying degrees of specialization for different tasks. The semantic memory loosely mimics the Neocortex as it constitutes a parametric memory system that builds structural representations for generalization that interacts with the instance level episodic memory (Section \ref{sec:formulation}) to efficiently accumulate and consolidate knowledge.

\subsection{Synaptic Consolidation} \label{sec:synap_cons}
In the brain, synaptic consolidation supports CL by reducing the plasticity of synapses that are considered important to previously learned tasks. This is enabled by a complex molecular machinery in biological synapses that allows multiples scales of learning by mediating the plasticity of individual synapses \citep{redondo2011making}. In DNNs, however, weights are only described by a scalar value and employing synaptic consolidation requires an additional estimate of the importance of synapses for the previous tasks. Thus, we utilize the Fisher information matrix (FIM) to estimate the importance of each parameter \citep{kirkpatrick2017overcoming}. We maintain a running approximation of FIM, similar to the online variants of EWC~\citep{chaudhry2018riemannian,schwarz2018progress}, but make several modifications to make the algorithm suitable for General-IL and enable effective consolidation of information across the tasks:
\begin{enumerate}[i.]
\setlength{\itemindent}{0em}
\itemsep0em
    \item We do not utilize the task boundaries and instead update the running estimate of FIM throughout the training trajectory stochastically using an update probability.
    \item Instead of evaluating the FIM on the training data of the current task, we use the samples in the memory buffer as they approximate the joint distribution of all the seen tasks. This provides a better estimate of the importance of a parameter for all the tasks and avoids the update from biasing the running estimate towards the current task.
    \item As the semantic memory effectively consolidates information across the tasks, and provides a better estimate of the optimal parameters for the joint distribution of tasks seen so far (Figure \ref{fig:working_semantic} in Appendix), we use it to evaluate the FIM instead of the working model. 
    \item Subsequently, the working model weights are anchored to the important parameters of the semantic memory. Since the synaptic regularization term acts as a proxy for the gradients of the previous tasks~\cite{hadsell2020embracing}, the consolidated weights of the semantic memory can provide a better estimate and more optimal constraints for the working model update. The use of semantic memory as anchor parameters has the additional benefits of not requiring to save the model's state which saves memory and not over constraining the model by anchoring it to parameters at the task boundary which may be unreliable and may hamper consolidation of information. 
    \item Finally, we note that one shortcoming of the regularization-based approaches is that they do not consider the structural functionality of filters in the CNNs. Each filter consists of a unit that extracts a specific feature from the input. Therefore, allowing large changes in some parameters while penalizing changes in others fails to prevent a drift in the functionality of the filter unit which might be important for the previous tasks. For instance, a horizontal edge detector requires each parameter of the filter to have certain values and changes in even a few parameters changes the functionality of the filter. To address this, we modify the FIM such that a uniform penalty is applied to all parameters of a filter. Specifically, we adjust the FIM for penalizing the weight changes such that the importance of each filter in the convolutional layer is given by the mean importance value of its parameters.
\end{enumerate}

\begin{algorithm*}[t]
\caption{SYNERgy algorithm for general continual learning}
\label{algo:SYNERgy}
\begin{algorithmic}[1]
\Statex {\bf Input:} {data stream $\mathcal{D}$; learning rate $\eta$; penalty terms $\lambda$, $\beta$; update probabilities $r_s$, $r_F$; decay parameters $\alpha_S$, $\alpha_F$}

\Statex {\bf Initialize: }{$\theta_s = \theta_w$}
\Statex {}{$\mathcal{M}\xleftarrow{}\{\}$} 

\While{Training}
\State Sample data: {$(x_b,y_b) \sim \mathcal{D}$} and {$(x_m,y_m) \sim \mathcal{M}$} 

\State {Calculate adjusted Fisher information matrix: $\mathcal{F}_{adj}$}
\State {Calculate loss: $\mathcal{L} = \mathcal{L}_{sl} + \lambda \mathcal{L}_{sr} +  \beta \mathcal{L}_{sc}$  ~~~~~(Eqs. \ref{eq:sl}, \ref{eq:sr} \& \ref{eq:sc})} 

\State {Update working model: $\theta_w \xleftarrow{} \theta_w - \eta \nabla_{\theta_w}\mathcal{L}$}

\State {Update semantic memory: $\theta_s \leftarrow \alpha_s\theta_s+(1-\alpha_s)~\theta_w,~~~~~if~~ r_s >a \sim U(0,1)$ ~~~~~(Eq. \ref{eq:sem_update})}
\State {Update Fisher information matrix: $\mathcal{F}^*_{\theta_s} \leftarrow \alpha_F\mathcal{F}^*_{\theta_s}+(1-\alpha_F)~\mathcal{F}_{\theta_s},~~~~~if~~ r_F >a \sim U(0,1)$ ~~~~~(Eqs. \ref{eq:Fisher_mat} \& \ref{eq:fish_update})}

\State {Update episodic memory: $\mathcal{M} \xleftarrow{} \text{Reservoir} (\mathcal{M}, (x_b,y_b))$}
\EndWhile
\Statex \Return{$\theta_s$}
\end{algorithmic}
\end{algorithm*}

\subsection{Formulation} \label{sec:formulation}
SYNERgy employs a dual memory experience replay mechanism in conjunction with synaptic consolidation. It maintains a fixed episodic memory $\mathcal{M}$ with budget size $\mathcal{B}$ through \textit{Reservoir sampling} and an additional semantic memory $\theta_s$ which aggregates the weights of the working model $\theta_w$ by taking an exponential moving average using Eq. \ref{eq:sem_update}. While training, samples from the current task, $(x_b,y_b) \sim \mathcal{D}_t$, are interleaved with the memory buffer samples, $(x_m,y_m) \sim \mathcal{M}$. The overall loss function for updating the working model comprises of three components as follows:

\textbf{Supervised Loss:} consists of learning from the ground truth labels of the incoming stream and the memory buffer samples through the task loss, $\mathcal{L}_{t}$, and episodic replay loss, $\mathcal{L}_{er}$. This ensures that the model learns from the approximate joint distribution of all the tasks seen so far. We use the standard Cross-Entropy loss for object recognition for both the task and episodic replay losses.
\begin{equation} \label{eq:sl}
    \mathcal{L}_{sl} =  \mathcal{L}_{t}(f(x_b;\theta_w, y_b)) + \mathcal{L}_{er}(f(x_m;\theta_w, y_m))
\end{equation}
\begin{equation} \label{eq:ce}
    \mathcal{L}_{ce} =  \mathcal{L}_{t}(f(x_b;\theta_w, y_b)) + \mathcal{L}_{er}(f(x_m;\theta_w, y_m))
\end{equation}

\textbf{Semantic Replay:} 
To provide additional structural information about the previous tasks, we utilize the semantic memory to extract the consolidated logits for effective replay of the memory samples. As the semantic memory aggregates information across the tasks, it takes into account the required adaptations in the decision boundary and the feature space to provide consolidated logits for replay, which facilitates the consolidation of knowledge~\citep{arani2022learning}.
Therefore, we apply semantic replay between the logits of the working model and the semantic memory on the memory samples, which encourages the model to maintain the similarity structure between the classes.
\begin{equation} \label{eq:sr}
    \mathcal{L}_{sr} =  \mathbb{E}_{x_m \sim M}[(f(x_m; \theta_w)-f(x_m; \theta_s))^2].
\end{equation}

\textbf{Synaptic Consolidation:}
To incorporate synaptic consolidation, we utilize the empirical FIM, $\mathcal{F}_{\theta_s}$, to estimate the importance of each parameter for the encountered tasks. 
At task t, we use the semantic memory to evaluate the FIM on all the samples in the memory buffer to approximate the expectation over the joint distribution of tasks seen so far, $\cup_{i=1}^{t}\mathcal{D}_i$. 
Consequently, the semantic memory is used as an anchor for constraining the network update. Our approach provides a better estimate of the parameter's importance as the semantic memory maintains consolidated information across the tasks and the Reservoir sampling in the episodic memory approximates the joint distribution of tasks,
\begin{equation} \label{eq:Fisher_mat}
    \mathcal{F}_{\theta_s} = \mathbb{E}_{(x,y) \sim \mathcal{M}} \Big[\Big(\frac{\partial \log p(y|x;{\theta_s})}{\partial \theta_s}\Big) \Big(\frac{\partial \log p(y|x;{\theta_s}) }{\partial \theta_s}\Big)^T\Big].
\end{equation}

As our focus is on General-IL, we avoid utilizing the task boundaries to calculate the FIM and save the model's state as anchor parameters. Instead, we take a stochastic approach to evaluate the FIM periodically throughout the training trajectory and aggregate them using an exponential moving average,
\begin{equation} \label{eq:fish_update}
    \mathcal{F}^*_{\theta_s} \leftarrow \alpha_F\mathcal{F}^*_{\theta_s}+(1-\alpha_F)~\mathcal{F}_{\theta_s},~~~~~if~~ r_F>a \sim U(0,1)
\end{equation}
where $\alpha_F$ is the decay parameter to control the strength of update and $ r_F$ is the update rate.

Furthermore, to maintain the structural functionality of filters in CNNs as noted in Section \ref{sec:synap_cons}, we adjust the importance estimate such that importance is defined at a filter level instead of individual parameters. We use the mean of the filter parameters in the aggregated FIM, $\mathcal{F}^*_{\theta_s}$, to estimate the importance of a filter (the resulting adjusted FIM is denoted by $\mathcal{F}_{adj}$). The synaptic consolidation loss is then given by the quadratic penalty weighted by $\mathcal{F}_{adj}$,
\begin{equation} \label{eq:sc}
    \mathcal{L}_{sc} = \mid\mid \theta_w - \theta_s \mid\mid^2_{\mathcal{F}_{adj}}.
\end{equation}
\textbf{Overall Loss:}
The working model is then updated by the weighted sum of individual losses, 
\begin{equation} \label{eq:loss}
    \mathcal{L} = \mathcal{L}_{sl} + \lambda \mathcal{L}_{sr} +  \beta \mathcal{L}_{sc}
\end{equation}
where $\lambda$ and $\beta$ control the strength of semantic replay and synaptic consolidation losses, respectively. Further details of the method are given in Algorithm \ref{algo:SYNERgy}. Note that for inference, we use the semantic memory as it gradually consolidates structural representations for generalization across the tasks (see Figure \ref{fig:task_perf}).

\begin{table*}[tb]
\caption{Comparison with prior works on various CL settings. The baseline results for Class-IL and Domain-IL on 200 and 500 buffer sizes are from \citet{buzzega2020dark} and MNIST-360 results are from \citet{arani2022learning}. 
}
\label{tab:all-datasets}
\centering
\begin{tabular}{@{\extracolsep{4pt}}llccccccc@{}}
\toprule
\multirow{2}{*}{{Buffer}} & \multirow{2}{*}{{Method}} & \multicolumn{2}{c}{\textbf{Class-IL}} & \textbf{Domain-IL} & \multicolumn{3}{c}{\textbf{General-IL}} \\ \cmidrule{3-4} \cmidrule{5-5} \cmidrule{6-8}
 &  &  {S-CIFAR10} & {S-TinyImg} & {R-MNIST} & {MNIST360} & {GCIL-U} & {GCIL-L} \\ \midrule
 
\multirow{2}{*}{–} & JOINT & 92.20\tiny±0.15 & 59.99\tiny±0.19 & 95.76\tiny±0.04 & 82.98\tiny±3.24 & 58.36\tiny±1.02 & 56.94\tiny±1.56 \\
 & SGD & 19.62\tiny±0.05 & 7.92\tiny±0.26 & 67.66\tiny±8.53 & 19.09\tiny±0.69 & 12.67\tiny±0.24 & 22.88\tiny±0.53 \\  \midrule
 
\multirow{3}{*}{–} & oEWC & 19.49\tiny±0.12 & 7.58\tiny±0.10 & 77.35\tiny±5.77 & - & - & - \\
 & SI   & 19.48\tiny±0.17 & 6.58\tiny±0.31 & 71.91\tiny±5.83 & - & - & - \\
 & LwF  & 19.61\tiny±0.05 & 8.46\tiny±0.22 & - & - & - & - \\ \midrule

\multirow{4}{*}{200} & ER & 44.79\tiny±1.86 & 8.49\tiny±0.16 & 85.01\tiny±1.90 & 49.27\tiny±2.25 & 16.40\tiny±0.37 & 19.27\tiny±0.77 \\ 
 & DER++& 64.88\tiny±1.17 & 10.96\tiny±1.17 & 90.43\tiny±1.87& 54.16\tiny±3.02 & 18.84\tiny±0.60 & 26.94\tiny±1.27 \\ 
 & CLS-ER & 66.19\tiny±0.75 & \textbf{21.95}\tiny±0.26 & \textbf{92.26}\tiny±0.18 & \textbf{66.37}\tiny±0.83 & 25.06\tiny±0.81 & 28.54\tiny±0.87 \\
 & SYNERgy & \textbf{66.27}\tiny±1.49 & 21.08\tiny±0.98 & 92.02\tiny±0.83 & 65.23\tiny±0.52 & \textbf{30.00}\tiny±2.63 & \textbf{32.24}\tiny±2.14 \\ \midrule

\multirow{4}{*}{500} & ER & 57.74\tiny±0.27 & 9.99\tiny±0.29 & 88.91\tiny±1.44 & 65.04\tiny±1.53 & 28.21\tiny±0.69 & 20.30\tiny±0.63 \\ 
 & DER++  & 72.70\tiny±1.36 & 19.38\tiny±1.41 & 92.77\tiny±1.05 & 69.62\tiny±1.59 & 32.92\tiny±0.74 & 25.82\tiny±0.83 \\ 
 & CLS-ER & \textbf{75.22}\tiny±0.71 & 29.61\tiny±0.54 & \textbf{94.06}\tiny±0.07 & 75.70\tiny±0.41 & 36.34\tiny±0.59 & 28.63\tiny±0.68 \\
 & SYNERgy  & 73.85\tiny±0.73  & \textbf{31.26}\tiny±0.67 & 93.37\tiny±0.45 & \textbf{76.48}\tiny±0.83 & \textbf{38.45}\tiny±0.46 & \textbf{33.07}\tiny±2.28 \\ \midrule

\multirow{4}{*}{1000} & ER & 70.64\tiny±1.13 & 13.23\tiny±0.39 & 89.44\tiny±1.34 & 75.18\tiny±1.50 & 31.98\tiny±0.72 & 34.13\tiny±0.83 \\ 
 & DER++ & 77.42\tiny±1.06 & 24.76\tiny±1.21 & 93.72\tiny±0.56 & 76.03\tiny±1.61 & 38.95\tiny±0.56 & 33.64\tiny±0.88 \\ 
 & CLS-ER  & 76.44\tiny±0.44 & 34.09\tiny±0.37 & \textbf{93.73}\tiny±0.20 & 79.54\tiny±0.34 & 39.69\tiny±0.66 & 39.52\tiny±0.91 \\
 & SYNERgy & \textbf{77.47}\tiny±1.15 & \textbf{36.49}\tiny±0.58 & 93.63\tiny±0.49 & \textbf{80.96}\tiny±0.33 &  \textbf{41.08}\tiny±0.59 & \textbf{39.96}\tiny±1.59 \\ 
\bottomrule
\end{tabular}
\end{table*}

\section{Empirical Evaluation}
We compare SYNERgy with regularization- (oEWC, SI), rehearsal- (ER, DER++), knowledge distillation-(LwF;~\cite{li2017learning}) and multiple memory-based (CLS-ER) methods under uniform experimental settings on varying CL scenarios (details of CL protocols and experimental setup are provided in Appendix \ref{eval-protocols} and \ref{exp-setup}). The lower bound is provided by standard training without any measure to avoid forgetting (SGD) and the upper bound is given by the joint training of all tasks (JOINT).  

Table \ref{tab:all-datasets} shows the performance of the models on different datasets with varying complexities in the Class-IL, Domain-IL and General-IL settings. SYNERgy performs on par with the state-of-the-art rehearsal based approaches or provides generalization gains in almost all the considered scenarios. Particularly in the challenging Class-IL setting with low buffer size and higher data complexity, SYNERgy improves the performance gains. The considerable performance gap between the regularization- and rehearsal-based approaches present a compelling case for the necessity of a small memory buffer to learn the joint distribution of the sequential tasks. This can be attributed to modeling the weight importance based on the local information at the task boundaries which can be unreliable and over-restrictive for constraining the model update. Furthermore, they fail to account for the required consolidation of information across the tasks and modifications in the representation space to discriminate across all the tasks and not only among each task. Synaptic consolidation in our method addresses these shortcomings.

On the other hand, the gap between the baseline rehearsal approach ER and our method in the more practical lower buffer regime (200 and 500) shows that merely intertwining the memory samples and current task samples is not sufficient to learn the joint distribution. The network can benefit both from extracting and replaying more information (e.g. neural activations) from the previous state as well as restricting the mode update to avoid overwriting important parameters. DER++ addresses the former by saving the soft-targets (logits) alongside the samples in memory and the performance improvement shows the effectiveness of providing additional information about the similarity structure of the classes. However, the local logits from the previous state of the model in their method fail to account for the adaptation of the decision boundaries and representation space required to distinguish classes in previous tasks from the new classes. Therefore, enforcing the model to match the sub-optimal logits and thereby a similarity structure that may not hold hampers the consolidation of information. This is more pronounced in the challenging S-TinyImageNet (referred to as S-TinyImg) setting where both the number of classes and inter-task similarity is high. In contrast, the aggregating and consolidating information across the tasks in the semantic memory provides a global context and an updated similarity structure for replay in the logits. The performance of CLS-ER and our method highlights the benefits of multiple memory systems in CL. While their method benefits from maintaining two semantic memories, it lacks a mechanism for synaptic consolidation which our method addresses. The task performance analysis in Section \ref{sec:char} shows that SYNERgy provides a better balance between the plasticity and stability of the model.

We also compare the performance of the models suitable for General-IL on the challenging MNIST360 and GCIL datasets.
SYNERgy provides significant performance improvement in both of these settings where the model needs to effectively accumulate and transfer knowledge of a class across repeated exposure and deal with multiple distribution shifts. The sub-optimal logits in DER++ fail to take advantage of the improved representations from the additional samples of the recurring classes in later tasks whereas the semantic memory consolidates information across different recurrences. The performance improvement over CLS-ER in this challenging setting, highlights the benefit of incorporating synaptic consolidation in the muliple memory systems for enabling general incremental learning. SYNERgy effectively employs a mechanism for synaptic consolidation that is suitable for the blurry task boundaries in General-IL which was missing in literature, and opens up a new direction of research that incorporates both these mechanisms which are critical components of the learning machinery of the brain into DNNs to enable effective CL. 

\begin{figure}[t]
 \centering
 \includegraphics[width=.98\textwidth]{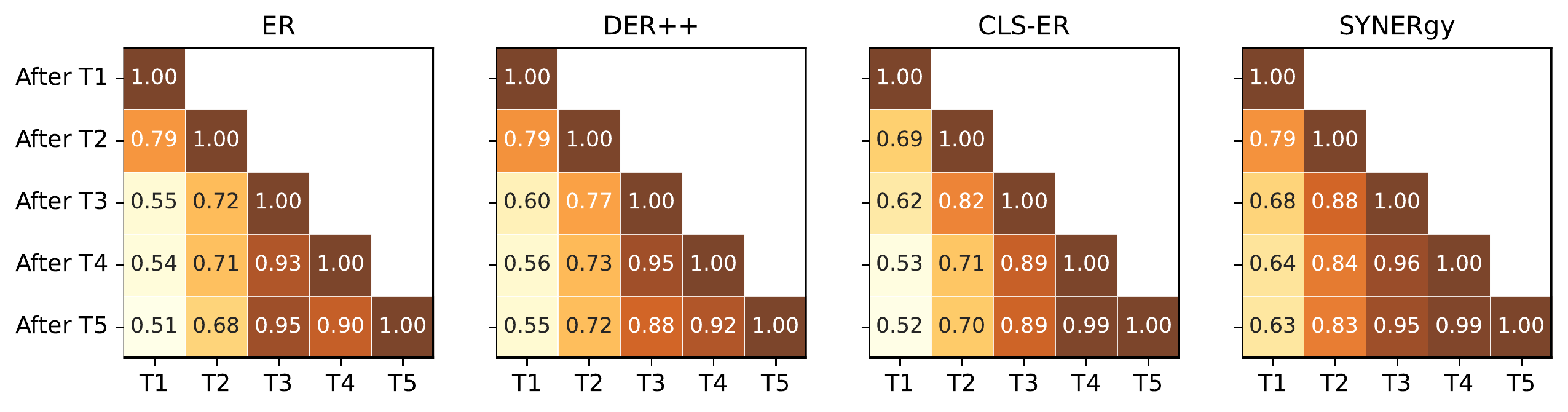}
 \caption{Drift in the learned parameters as training progresses. We report the average cosine similarity between the layers of task models (x-axis) at the end of each task (y-axis). Synaptic consolidation enables the model to consolidate knowledge while keeping the parameters closer to their optimal task parameters.}
 \label{fig:weight}
\end{figure}

\begin{figure}[t]
 \centering
 \includegraphics[width=.95\textwidth]{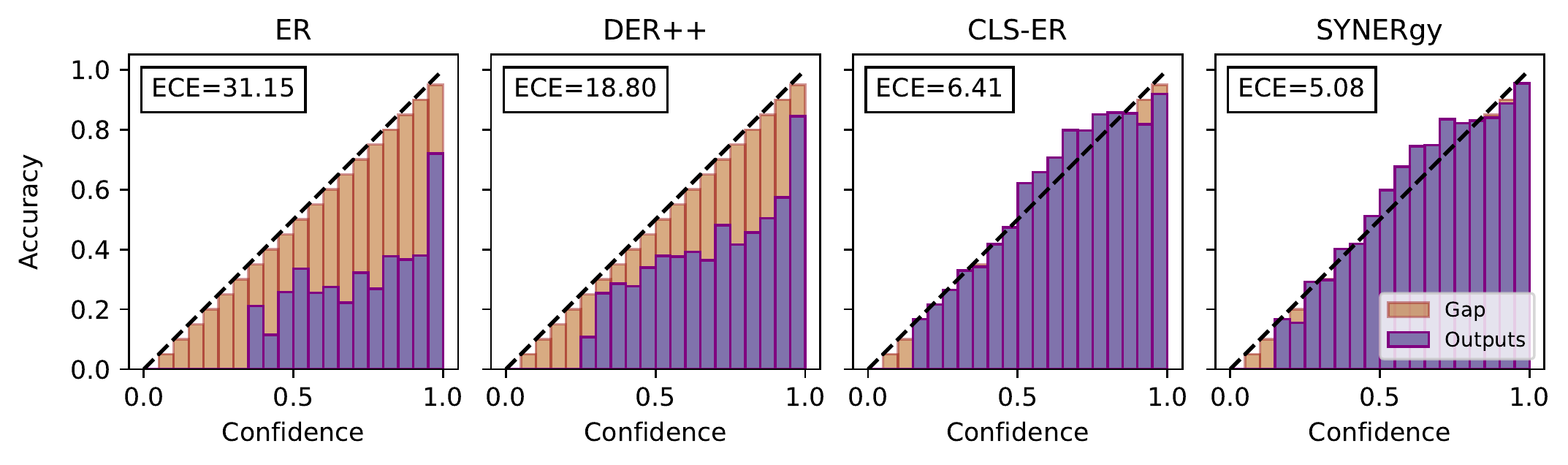}
 \caption{Reliability plots and expected calibration error: SYNERgy improves the reliabiliy of predictions even with lower buffer size. See Figures \ref{fig:calib-cifar} and \ref{fig:calib-tinyimg} for other dataset and buffer sizes.}
 \label{fig:calib}
\end{figure}


\section{Model Characteristics} \label{sec:char}
Here, we probe into the characteristics distilled in the model trained on S-CIFAR-100 with 500 buffer size with multiple memory-based replay and synaptic consolidation and compare it with ER, DER++, and CLS-ER. 

\textbf{Drift in Parameters:} An important aspect of enabling effective CL is to avoid the drift in the optimal parameters of the model for previous tasks to mitigate forgetting. To analyze the impact of synaptic consolidation in SYNERgy, we evaluate the change in the learned parameters at the end of each task as training progresses. We evaluate the similarity between the parameters by averaging the cosine similarity between the layers of the model. To account for the scale differences and linear changes, we normalize the weights of each layer by dividing by the maximum weight. Figure \ref{fig:weight} shows that SYNERgy effectively reduces the changes in the parameters. In particular, the final parameters of SYNERgy (T5) are considerably closer to the earliest tasks (T1 and T2) compared to the baselines. This suggests that our method effectively employs synaptic consolidation to reduce forgetting by keeping the parameters closer to the optimal parameters of the previous tasks.

\begin{figure}[t]
    \centering
    \includegraphics[width=.98\textwidth]{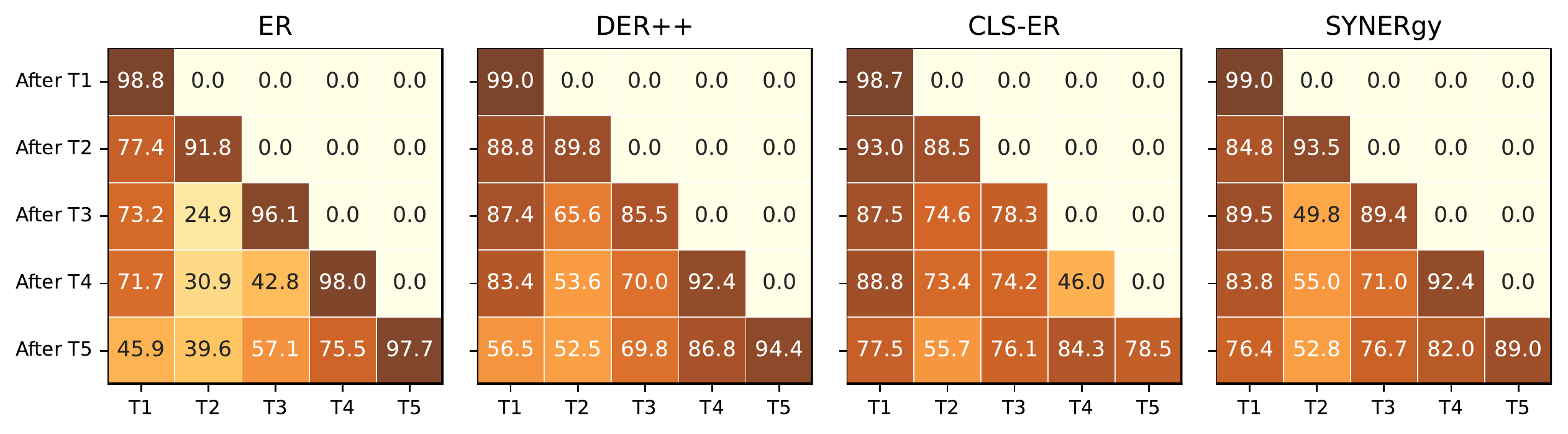}
    \caption{Task-wise performance of methods: Heatmap shows the performance of tasks (x-axis) evaluated at the end of training each task in the sequence (y-axis). The consolidated structural representation in SYNERgy generalizes well across the tasks and significantly reduces forgetting.}
\label{fig:task_perf}
\end{figure}

\textbf{Model Calibration} measures how indicative the probability of predictions are of their accuracies. DNNs tend to be poorly calibrated which reduces the reliability of their predictions. Combined with the bias towards recent tasks, this issue is even more pronounced in CL. Following \citet{guo2017calibration}, we plot accuracy as a function of confidence (reliability plot) and calculate the weighted average over the absolute accuracy/confidence difference (Expected Calibration Error). Figure \ref{fig:calib} shows that SYNERgy improves the calibration of model, hence the reliability of its predictions.

\textbf{Task Performances:}
The average accuracy across the tasks alone fail to evaluate how well a CL method is able to maintain a balance between avoiding forgetting (stability) and allowing the model to learn a new task (plasticity). To this end, we evaluate the test set task-wise performance at the end of training of new tasks. The difference between the performance of the model when it was trained on the task and at the end of training shows the degree of forgetting whereas the performance of the model when it was trained shows how well the model is able to learn a new task. Figure \ref{fig:task_perf} shows that SYNERgy provides a better balance between stability and plasticity compared to the baselines even with lower buffer size. While CLS-ER is able to reduce forgetting, the performance of the model on the current task suffers perhaps due to the lower rate of updating the stable model in the method. On Task 5, SYNERgy provides 89\% accuracy compared to 78\% in CLS-ER while retaining comparable performance on previous tasks. This suggests that synaptic consolidation in conjunction with dual memory experience replay provides a viable approach for tackling the stability and plasticity dilemma in CL.

\begin{figure}[t]
    \centering
    \begin{minipage}{0.525\textwidth}
        \begin{figure}[H]
        \centering
        \includegraphics[width=1\textwidth]{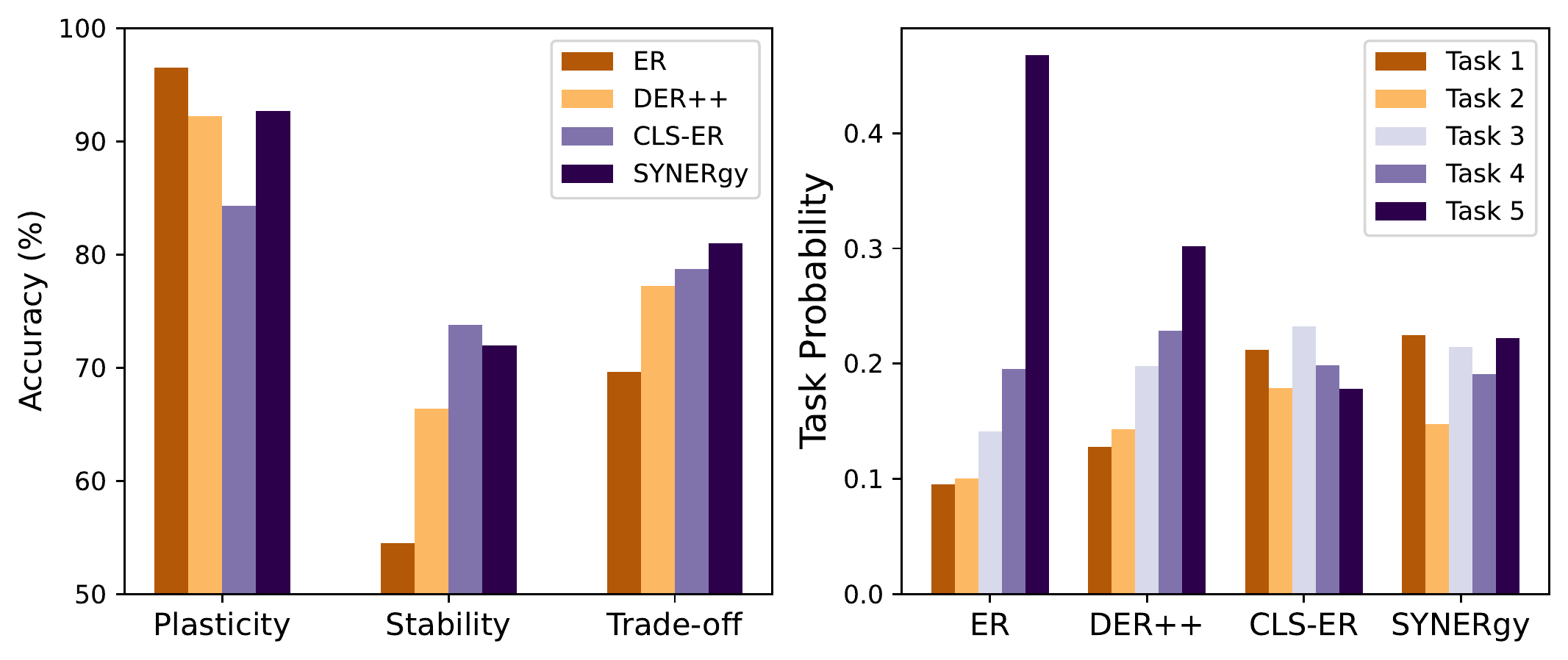} \\ 
        a) Stability-Plasticity Trade-off~~~~~~~~~~b) Task probability
        \caption{a) SYNERgy provides a better trade-off between the stability and plasticity of the model. b) It also reduces the bias towards recent tasks and provides more uniform predictions across the tasks. See Figures \ref{fig:task_prob-cifar} and \ref{fig:task_prob-tinyimg} for other buffer sizes and datasets.}
        \label{fig:tradeoff_taskProb}
        \end{figure}
    \end{minipage}%
    \hspace{0.015\textwidth}%
        \begin{minipage}{0.45\textwidth}
        \begin{table}[H]
            \centering
            \caption{\textbf{Ablation Study:} DM and TB denote the usage of dual memories and task boundary information for training, respectively. SC indicates which samples ($\mathcal{M}$ or $\mathcal{D}_t$) and model ($\theta_s$ or $\theta_w$) are utilized to evaluate FIM and whether it is adjusted $\mathcal{F}^{adj}$. All the components contribute to the performance gains.}
            \label{tab:abl}
            \begingroup
            \setlength{\tabcolsep}{3pt}
            \renewcommand{\arraystretch}{1.2}
            \begin{tabular}{lcccc}
            \toprule
             & DM & TB & SC & Accuracy \\ \midrule
            SYNERgy & \cmark & \xmark & $\mathcal{M},\mathcal{F}^{adj}_{\theta_s}$ & {\bf 73.85}\tiny±0.73 \\
            $-\mathcal{F}_{adj}$ & \cmark & \xmark & $\mathcal{M},\mathcal{F}^*_{\theta_s}$  & 72.88\tiny±1.63 \\
            $-\mathcal{M},\theta_S$ & \cmark & \xmark & $\mathcal{D}_t,\mathcal{F}^{adj}_{\theta_w}$ & 72.26\tiny±0.82 \\
            Mean-ER & \cmark & \xmark & \xmark & 71.99\tiny±1.26 \\
             + oEWC & \cmark & \cmark & $\mathcal{D}_t,\mathcal{F}^*_{\theta_w}$ & 67.73\tiny±1.37 \\ 
            ER & \xmark & \xmark & \xmark & 57.74\tiny±0.27\\
            \bottomrule
            \end{tabular}
            \endgroup
        \end{table}
    \end{minipage}
    
\end{figure}

\textbf{Task Probabilities:} The sequential learning of tasks in CL biases the model towards the current task~\citep{wu2019large} which affects performance on earlier tasks considerably. To test whether synaptic consolidation in SYNERgy reduces the bias towards recent tasks, we evaluate the probability of predicting each task at the end of training. We take the softmax output of each sample in the test set and average the probabilities of the classes in each task. Figure \ref{fig:tradeoff_taskProb}b shows that while ER and DER++ show considerable bias towards the last task when trained with low buffer size, the multiple memories in CLS-ER and SYNERgy provide more uniform prediction probabilities across the tasks. SYNERgy. CLS-ER reduces the probability of predicting the last task which affects the performance of the model on recent tasks. SYNERgy, on the other hand, leads to more uniform task probabilities without compromising on the recent task.

\textbf{Stability-Plasticity Trade-off:}
To dissect how well SYNERgy tackles the stability-plasticity dilemma that lies at the core of CL, we take a closer look at the task-wise performance matrix, $\mathcal{T}$, where $\mathcal{T}_{i,j}$ corresponds to the test accuracy of task $j$ after learning task $i$. At a given task, $t$, in sequence, we consider \textit{stability}, $\mathcal{S}$, as the average performance of all the previous $t-1$ tasks after learning task $t$, $mean(\mathcal{T}_{t, 1:t-1}$). \textit{Plasticity}, $\mathcal{P}$, is measured by the average performance of the tasks when they are first learned, $mean(Diag(\mathcal{T}))$. Hence, \textit{stability} measures how well the model avoids forgetting and retains performance on the previous tasks whereas \textit{plasticity} measures the capability of the model to learn the new task. Effective CL requires an optimal balance between stability and plasticity. However, there isn't a standardized way of measuring this trade-off. We, therefore, propose a trade-off measure that provides a formal way of comparing how well different methods maintain the balance. The \textit{Trade-off} is measured as follow:
\begin{equation} \label{eq:tradeoff}
    \textit{Trade-off} = \frac{2\times\mathcal{S}\times\mathcal{P}}{\mathcal{S} + \mathcal{P}}
\end{equation}
Figure \ref{fig:tradeoff_taskProb}a compares the stability, plasticity and trade-off of the different models evaluated at the end of training all $T$ tasks in sequence. SYNERgy maintains a better trade-off compared to the baselines. CLS-ER provides higher stability largely due to the lower frequency of updates for the stable model but it comes at the cost of plasticity. SYNERgy, on the other hand, not only provides comparable plasticity to DER++ but also considerably higher stability.

\textbf{Ablation Study:}
To gain insights into the effect of the different components of SYNERgy, we conduct a series of ablation experiments. We systematically remove the components and show their effect on the performance of the models trained in Table \ref{tab:abl}. We can derive several insights from the results. Both synaptic consolidation and dual memory experience replay play a critical role in reducing forgetting.
Mean-ER provides a significant performance improvement over typical ER. Our proposed Synaptic consolidation in SYNERgy provides further performance gain and improves the stability of the model. Furthermore, evaluating the Fisher information matrix using the semantic memory on the memory buffer instead of the working model on the current task training data provides a better estimate of the importance of parameters for all encountered tasks. It also shows that not accounting for the functional integrity of the convolution filter (-$\mathcal{F}_{adj}$) reduces performance. Our results suggest that more specialized algorithms for calculating the parameter importance at the filter level instead of post-processing may further improve regularization.
Importantly, merely plugging in regularization in Mean-ER (+oEWC) not only adds the reliance on task boundaries but is also detrimental to learning. oEWC anchors the weights of the model to their state at the task boundary which over-constrains the model update and hampers knowledge consolidation. Furthermore, the importance estimates are biased towards the current task. Our proposed modifications, in addition to making the method suitable for general continual learning, enable knowledge consolidation by maintaining an updated estimate of parameter importance across the encountered tasks and anchoring the weights to the consolidated weights in the semantic memory. Also, our method does not require additional memory to save the reference weights. Our study, therefore, not only highlights the benefits of incorporating both experience replay and regularization in a dual-memory architecture but also emphasizes the importance of design choices to enable consolidation as the de-facto combination is detrimental to knowledge consolidation.

\section{Conclusion}
Inspired by the mechanisms employed by the brain to enable efficient consolidation of information and protection from the erasure of past memories, we proposed a novel dual memory replay mechanism in synergy with synaptic consolidation for general incremental learning. Our proposed method, SYNERgy, maintains an additional semantic memory that interacts with the episodic memory for effective semantic replay to consolidate information across the tasks. Moreover, synaptic consolidation is employed by tracking the importance of parameters throughout the learning trajectory and penalizing changes in parameters that are considered important for previous tasks. We proposed several modifications to make the regularization required for synaptic consolidation suitable for general incremental learning such that the method does not utilize the task boundaries at training and inference. 
We showed strong empirical results on various challenging CL settings and demonstrate the effectiveness of our method in providing a better balance between stability and plasticity of the model. Furthermore, SYNERgy improves the reliability of the model, reduces the drift in the parameters of the previous tasks and alleviates the bias towards recent tasks. Our results show the potential of employing synaptic consolidation in conjunction with experience replay in a dual memory system and motivates further work in this promising direction.


\bibliography{egbib}

\begin{thebibliography}{35}
\providecommand{\natexlab}[1]{#1}
\providecommand{\url}[1]{\texttt{#1}}
\expandafter\ifx\csname urlstyle\endcsname\relax
  \providecommand{\doi}[1]{doi: #1}\else
  \providecommand{\doi}{doi: \begingroup \urlstyle{rm}\Url}\fi

\bibitem[Aljundi et~al.(2019)Aljundi, Lin, Goujaud, and
  Bengio]{aljundi2019gradient}
Rahaf Aljundi, Min Lin, Baptiste Goujaud, and Yoshua Bengio.
\newblock Gradient based sample selection for online continual learning.
\newblock In \emph{Advances in Neural Information Processing Systems}, pp.\
  11816--11825, 2019.

\bibitem[Arani et~al.(2022)Arani, Sarfraz, and Zonooz]{arani2022learning}
Elahe Arani, Fahad Sarfraz, and Bahram Zonooz.
\newblock Learning fast, learning slow: A general continual learning method
  based on complementary learning system.
\newblock In \emph{International Conference on Learning Representations}, 2022.
\newblock URL \url{https://openreview.net/forum?id=uxxFrDwrE7Y}.

\bibitem[Benna \& Fusi(2016)Benna and Fusi]{benna2016computational}
Marcus~K Benna and Stefano Fusi.
\newblock Computational principles of synaptic memory consolidation.
\newblock \emph{Nature neuroscience}, 19\penalty0 (12):\penalty0 1697--1706,
  2016.

\bibitem[Buzzega et~al.(2020)Buzzega, Boschini, Porrello, Abati, and
  Calderara]{buzzega2020dark}
Pietro Buzzega, Matteo Boschini, Angelo Porrello, Davide Abati, and Simone
  Calderara.
\newblock Dark experience for general continual learning: a strong, simple
  baseline.
\newblock \emph{arXiv preprint arXiv:2004.07211}, 2020.

\bibitem[Chaudhry et~al.(2018{\natexlab{a}})Chaudhry, Dokania, Ajanthan, and
  Torr]{chaudhry2018riemannian}
Arslan Chaudhry, Puneet~K Dokania, Thalaiyasingam Ajanthan, and Philip~HS Torr.
\newblock Riemannian walk for incremental learning: Understanding forgetting
  and intransigence.
\newblock In \emph{Proceedings of the European Conference on Computer Vision
  (ECCV)}, pp.\  532--547, 2018{\natexlab{a}}.

\bibitem[Chaudhry et~al.(2018{\natexlab{b}})Chaudhry, Ranzato, Rohrbach, and
  Elhoseiny]{chaudhry2018efficient}
Arslan Chaudhry, Marc'Aurelio Ranzato, Marcus Rohrbach, and Mohamed Elhoseiny.
\newblock Efficient lifelong learning with a-gem.
\newblock \emph{arXiv preprint arXiv:1812.00420}, 2018{\natexlab{b}}.

\bibitem[Cichon \& Gan(2015)Cichon and Gan]{cichon2015branch}
Joseph Cichon and Wen-Biao Gan.
\newblock Branch-specific dendritic ca 2+ spikes cause persistent synaptic
  plasticity.
\newblock \emph{Nature}, 520\penalty0 (7546):\penalty0 180--185, 2015.

\bibitem[Clopath(2012)]{clopath2012synaptic}
Claudia Clopath.
\newblock Synaptic consolidation: an approach to long-term learning.
\newblock \emph{Cognitive neurodynamics}, 6\penalty0 (3):\penalty0 251--257,
  2012.

\bibitem[Farquhar \& Gal(2018)Farquhar and Gal]{farquhar2018towards}
Sebastian Farquhar and Yarin Gal.
\newblock Towards robust evaluations of continual learning.
\newblock \emph{arXiv preprint arXiv:1805.09733}, 2018.

\bibitem[Guo et~al.(2017)Guo, Pleiss, Sun, and Weinberger]{guo2017calibration}
Chuan Guo, Geoff Pleiss, Yu~Sun, and Kilian~Q Weinberger.
\newblock On calibration of modern neural networks.
\newblock In \emph{International Conference on Machine Learning}, pp.\
  1321--1330. PMLR, 2017.

\bibitem[Hadsell et~al.(2020)Hadsell, Rao, Rusu, and
  Pascanu]{hadsell2020embracing}
Raia Hadsell, Dushyant Rao, Andrei~A Rusu, and Razvan Pascanu.
\newblock Embracing change: Continual learning in deep neural networks.
\newblock \emph{Trends in cognitive sciences}, 24\penalty0 (12):\penalty0
  1028--1040, 2020.

\bibitem[Hassabis et~al.(2017)Hassabis, Kumaran, Summerfield, and
  Botvinick]{hassabis2017neuroscience}
Demis Hassabis, Dharshan Kumaran, Christopher Summerfield, and Matthew
  Botvinick.
\newblock Neuroscience-inspired artificial intelligence.
\newblock \emph{Neuron}, 95\penalty0 (2):\penalty0 245--258, 2017.

\bibitem[Hayes et~al.(2021)Hayes, Krishnan, Bazhenov, Siegelmann, Sejnowski,
  and Kanan]{hayes2021replay}
Tyler~L Hayes, Giri~P Krishnan, Maxim Bazhenov, Hava~T Siegelmann, Terrence~J
  Sejnowski, and Christopher Kanan.
\newblock Replay in deep learning: Current approaches and missing biological
  elements.
\newblock \emph{Neural Computation}, 33\penalty0 (11):\penalty0 2908--2950,
  2021.

\bibitem[He et~al.(2015)He, Zhang, Ren, and Sun]{he2015deep}
Kaiming He, Xiangyu Zhang, Shaoqing Ren, and Jian Sun.
\newblock Deep residual learning for image recognition. corr abs/1512.03385
  (2015), 2015.

\bibitem[Kirkpatrick et~al.(2017)Kirkpatrick, Pascanu, Rabinowitz, Veness,
  Desjardins, Rusu, Milan, Quan, Ramalho, Grabska-Barwinska,
  et~al.]{kirkpatrick2017overcoming}
James Kirkpatrick, Razvan Pascanu, Neil Rabinowitz, Joel Veness, Guillaume
  Desjardins, Andrei~A Rusu, Kieran Milan, John Quan, Tiago Ramalho, Agnieszka
  Grabska-Barwinska, et~al.
\newblock Overcoming catastrophic forgetting in neural networks.
\newblock \emph{Proceedings of the national academy of sciences}, 114\penalty0
  (13):\penalty0 3521--3526, 2017.

\bibitem[Krishnan et~al.(2019)Krishnan, Tadros, Ramyaa, and
  Bazhenov]{krishnan2019biologically}
Giri~P Krishnan, Timothy Tadros, Ramyaa Ramyaa, and Maxim Bazhenov.
\newblock Biologically inspired sleep algorithm for artificial neural networks.
\newblock \emph{arXiv preprint arXiv:1908.02240}, 2019.

\bibitem[Krizhevsky et~al.(2009)]{krizhevsky2009learning}
Alex Krizhevsky et~al.
\newblock Learning multiple layers of features from tiny images.
\newblock 2009.

\bibitem[Kumaran et~al.(2016)Kumaran, Hassabis, and
  McClelland]{kumaran2016learning}
Dharshan Kumaran, Demis Hassabis, and James~L McClelland.
\newblock What learning systems do intelligent agents need? complementary
  learning systems theory updated.
\newblock \emph{Trends in cognitive sciences}, 20\penalty0 (7):\penalty0
  512--534, 2016.

\bibitem[Li \& Hoiem(2017)Li and Hoiem]{li2017learning}
Zhizhong Li and Derek Hoiem.
\newblock Learning without forgetting.
\newblock \emph{IEEE transactions on pattern analysis and machine
  intelligence}, 40\penalty0 (12):\penalty0 2935--2947, 2017.

\bibitem[Lopez-Paz \& Ranzato(2017)Lopez-Paz and Ranzato]{lopez2017gradient}
David Lopez-Paz and Marc'Aurelio Ranzato.
\newblock Gradient episodic memory for continual learning.
\newblock In \emph{Advances in neural information processing systems}, pp.\
  6467--6476, 2017.

\bibitem[McCloskey \& Cohen(1989)McCloskey and
  Cohen]{mccloskey1989catastrophic}
Michael McCloskey and Neal~J Cohen.
\newblock Catastrophic interference in connectionist networks: The sequential
  learning problem.
\newblock In \emph{Psychology of learning and motivation}, volume~24, pp.\
  109--165. Elsevier, 1989.

\bibitem[Mi et~al.(2020)Mi, Kong, Lin, Yu, and Faltings]{mi2020generalized}
Fei Mi, Lingjing Kong, Tao Lin, Kaicheng Yu, and Boi Faltings.
\newblock Generalized class incremental learning.
\newblock In \emph{Proceedings of the IEEE/CVF Conference on Computer Vision
  and Pattern Recognition Workshops}, pp.\  240--241, 2020.

\bibitem[Parisi et~al.(2019)Parisi, Kemker, Part, Kanan, and
  Wermter]{parisi2019continual}
German~I Parisi, Ronald Kemker, Jose~L Part, Christopher Kanan, and Stefan
  Wermter.
\newblock Continual lifelong learning with neural networks: A review.
\newblock \emph{Neural Networks}, 113:\penalty0 54--71, 2019.

\bibitem[Pham et~al.(2021)Pham, Liu, and Hoi]{pham2021dualnet}
Quang Pham, Chenghao Liu, and Steven Hoi.
\newblock Dualnet: Continual learning, fast and slow.
\newblock \emph{Advances in Neural Information Processing Systems}, 34, 2021.

\bibitem[Pouransari \& Ghili(2015)Pouransari and Ghili]{pouransari2015tiny}
Hadi Pouransari and Saman Ghili.
\newblock Tiny imagenet visual recognition challenge.
\newblock \emph{CS231N course, Stanford Univ., Stanford, CA, USA}, 2015.

\bibitem[Redondo \& Morris(2011)Redondo and Morris]{redondo2011making}
Roger~L Redondo and Richard~GM Morris.
\newblock Making memories last: the synaptic tagging and capture hypothesis.
\newblock \emph{Nature Reviews Neuroscience}, 12\penalty0 (1):\penalty0 17--30,
  2011.

\bibitem[Riemer et~al.(2018)Riemer, Cases, Ajemian, Liu, Rish, Tu, and
  Tesauro]{riemer2018learning}
Matthew Riemer, Ignacio Cases, Robert Ajemian, Miao Liu, Irina Rish, Yuhai Tu,
  and Gerald Tesauro.
\newblock Learning to learn without forgetting by maximizing transfer and
  minimizing interference.
\newblock \emph{arXiv preprint arXiv:1810.11910}, 2018.

\bibitem[Ritter et~al.(2018)Ritter, Botev, and Barber]{ritter2018online}
Hippolyt Ritter, Aleksandar Botev, and David Barber.
\newblock Online structured laplace approximations for overcoming catastrophic
  forgetting.
\newblock In \emph{Advances in Neural Information Processing Systems}, pp.\
  3738--3748, 2018.

\bibitem[Schwarz et~al.(2018)Schwarz, Czarnecki, Luketina, Grabska-Barwinska,
  Teh, Pascanu, and Hadsell]{schwarz2018progress}
Jonathan Schwarz, Wojciech Czarnecki, Jelena Luketina, Agnieszka
  Grabska-Barwinska, Yee~Whye Teh, Razvan Pascanu, and Raia Hadsell.
\newblock Progress \& compress: A scalable framework for continual learning.
\newblock In \emph{International Conference on Machine Learning}, pp.\
  4528--4537. PMLR, 2018.

\bibitem[Tulving(2002)]{tulving2002episodic}
Endel Tulving.
\newblock Episodic memory: From mind to brain.
\newblock \emph{Annual review of psychology}, 53\penalty0 (1):\penalty0 1--25,
  2002.

\bibitem[Vitter(1985)]{vitter1985random}
Jeffrey~S Vitter.
\newblock Random sampling with a reservoir.
\newblock \emph{ACM Transactions on Mathematical Software (TOMS)}, 11\penalty0
  (1):\penalty0 37--57, 1985.

\bibitem[Wang et~al.(2022)Wang, Zhang, Ebrahimi, Sun, Zhang, Lee, Ren, Su,
  Perot, Dy, et~al.]{wang2022dualprompt}
Zifeng Wang, Zizhao Zhang, Sayna Ebrahimi, Ruoxi Sun, Han Zhang, Chen-Yu Lee,
  Xiaoqi Ren, Guolong Su, Vincent Perot, Jennifer Dy, et~al.
\newblock Dualprompt: Complementary prompting for rehearsal-free continual
  learning.
\newblock \emph{arXiv preprint arXiv:2204.04799}, 2022.

\bibitem[Wu et~al.(2019)Wu, Chen, Wang, Ye, Liu, Guo, and Fu]{wu2019large}
Yue Wu, Yinpeng Chen, Lijuan Wang, Yuancheng Ye, Zicheng Liu, Yandong Guo, and
  Yun Fu.
\newblock Large scale incremental learning.
\newblock In \emph{Proceedings of the IEEE/CVF Conference on Computer Vision
  and Pattern Recognition}, pp.\  374--382, 2019.

\bibitem[Yang et~al.(2009)Yang, Pan, and Gan]{yang2009stably}
Guang Yang, Feng Pan, and Wen-Biao Gan.
\newblock Stably maintained dendritic spines are associated with lifelong
  memories.
\newblock \emph{Nature}, 462\penalty0 (7275):\penalty0 920--924, 2009.

\bibitem[Zenke et~al.(2017)Zenke, Poole, and Ganguli]{zenke2017continual}
Friedemann Zenke, Ben Poole, and Surya Ganguli.
\newblock Continual learning through synaptic intelligence.
\newblock In \emph{International Conference on Machine Learning}, pp.\
  3987--3995. PMLR, 2017.

\end{thebibliography}
\bibliographystyle{collas2022_conference}

\newpage
\appendix
\newpage
\appendix

\setcounter{figure}{0}
\makeatletter 
\renewcommand{\thefigure}{S\@arabic\c@figure}
\makeatother
\setcounter{table}{0}
\makeatletter 
\renewcommand{\thetable}{S\@arabic\c@table}
\makeatother

\section{Evaluation Protocols} \label{eval-protocols}
To gauge the effectiveness and versatility of our method, we test it under varying CL scenarios, each of which presents its own set of challenges that the CL method needs to tackle.
In this study, we consider the following CL scenarios. 

\textbf{Class-IL} involves learning a new set of disjoint classes in subsequent tasks and the agent is required to distinguish among all the classes seen so far. Class-IL measures the ability of the agent to learn izable features, which requires consolidating acquired knowledge with new tasks to optimally adapt the representation space and decision boundaries. We consider CIFAR10~\citep{krizhevsky2009learning} and TinyImageNet~\citep{pouransari2015tiny} split into 5 and 10 tasks, respectively, and hence cover settings with varying datasets and task length complexities.   

\textbf{Domain-IL} evaluates the ability of the model to tackle changes in the input data distribution with the number of classes remaining fixed. It measures the ability of the model to learn features that are robust to the distribution shift. We consider R-MNIST~\citep{lopez2017gradient} which requires the agent to classify MNIST digits for 20 subsequent tasks with images rotated by a random angle between 0 and 180 degrees.

\textbf{General-IL} attempts to simulate the challenges in the real-world and involves training the method on longer sequences where the boundaries between the tasks are not distinct, the tasks themselves are not disjoint, and there are multiple sources of data distributions shifts~\citep{arani2022learning}. 
In addition to preventing catastrophic forgetting, the learning agent has to tackle several challenges: class imbalance, sample efficiency, learning an object over multiple recurrences and multiple distribution shifts.  We consider two General-IL settings which present different sets of challenges. MNIST360~\citep{buzzega2020dark} exposes the model to both a sharp distribution shift in classes and a smooth rotational distribution shift by presenting batches of two consecutive MNIST images (e.g. $\{0, 1\}$,
$\{1, 2\}$) and the samples are rotated at an increasing angle. GCIL~\citep{mi2020generalized} utilizes probabilistic modeling to sample the classes and data distributions from CIFAR100 dataset~\citep{krizhevsky2009learning} in each task. Hence, the number of classes in each task is not fixed, the classes can overlap and the sample size for each class can vary. GCIL contains a sequence of 20 tasks where each task has 1000 samples with a maximum of 50 classes and the sample distribution is either uniform (U) or longtail (L).

\section{Working Model Performance}
\begin{figure}[ht]
    \centering
    \begin{tabular}{cc}
         \includegraphics[width=.49\textwidth]{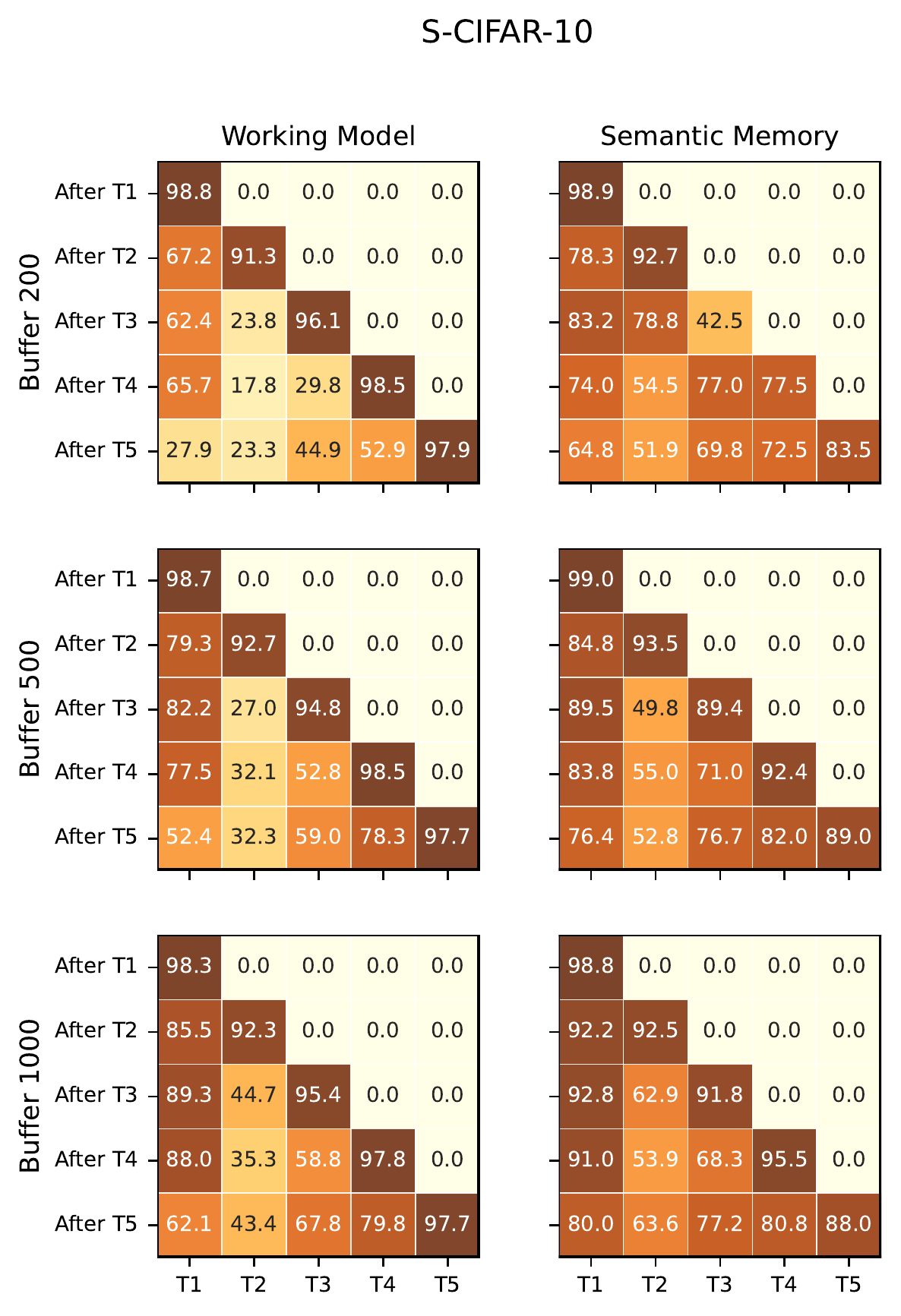} &
          \includegraphics[width=.49\textwidth]{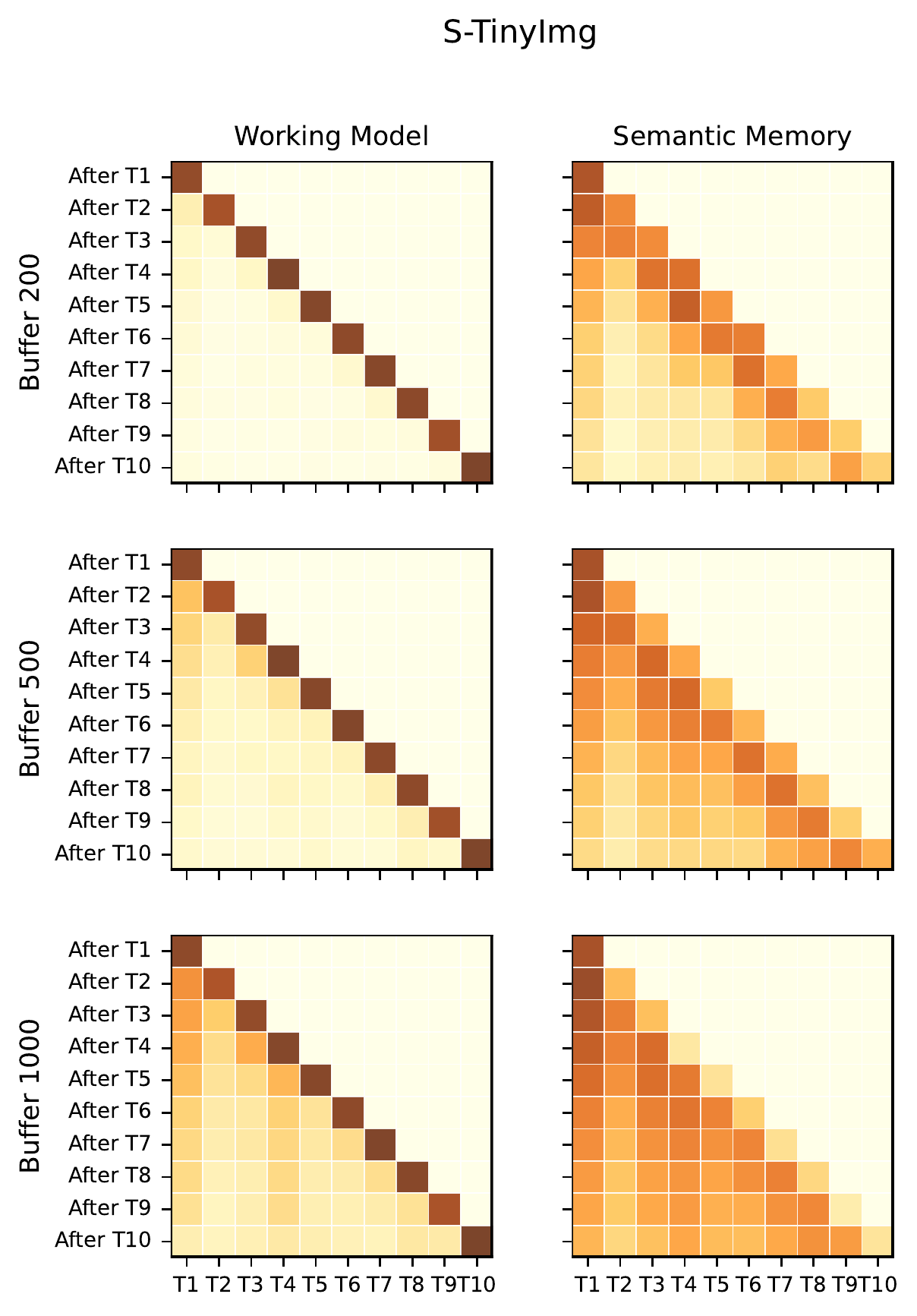}
    \end{tabular}
    \caption{Task-wise performance of working model and semantic memory evaluated on the test set of S-CIFAR10 and TinyImageNet datasets. Heatmap shows the performance of tasks (x-axis) evaluated at the end of training each task in the sequence (y-axis). Semantic memory buils consolidated structural representation which generalizes well across the tasks and is therefore suitable for inference.}
\label{fig:working_semantic}
\end{figure}

Here, we provide the performance of the working model on each of the evaluation setting and compare with the semantic memory. Table \ref{tab:working_perf} indicates that the semantic memory is able to effectively aggregate knowledge from the working model and build consolidated features for generalization across the tasks. Particularly, for the challenging data and memory-restrictive settings, the performance gap between the semantic memory and working model is considerable. Figure \ref{fig:working_semantic} further compares the task-wise performance and show that the semantic memory generalizes well across the tasks whereas the working model is more optimized for the recent task. These results highlight the effectiveness of maintaining semantic memory using the exponential moving average over the weights of the working model.

\begin{table}[ht]
\caption{Performance of the working model and semantic memory on the different evaluation settings. Semantic memory provides better generalization across all the considered setting especially in challenging dataset and memory restrictive scenarios.}
\label{tab:working_perf}
\centering
\begin{tabular}{l|lcc|lcc}
\toprule
\begin{tabular}[c]{@{}l@{}}Buffer\\ size\end{tabular} & Dataset & \begin{tabular}[c]{@{}c@{}}Semantic\\ memory\end{tabular} & \begin{tabular}[c]{@{}c@{}}Working \\ model\end{tabular} & Dataset & \begin{tabular}[c]{@{}c@{}}Semantic\\ memory\end{tabular} & \begin{tabular}[c]{@{}c@{}}Working \\ model\end{tabular} \\ \midrule
200 & \multirow{3}{*}{S-CIFAR-10} & 66.27\tiny±1.4 & 51.13\tiny±1.34 & \multirow{3}{*}{MNIST-360} & 65.23\tiny±0.52 & 59.84\tiny±0.87 \\
500 &  & 73.85\tiny±0.73 & 63.80\tiny±1.82 &  & 76.48\tiny±0.83 & 73.49\tiny±0.99 \\
1000 &  & 77.47\tiny±1.15 & 70.70\tiny±1.24 &  & 80.96\tiny±0.33 & 78.50\tiny±0.89 \\\midrule
200 & \multirow{3}{*}{S-TinyImg} & 21.68\tiny±1.51 & 10.46\tiny±0.33 & \multirow{3}{*}{GCIL-U} & 30.00\tiny±2.63 & 25.27\tiny±2.82 \\
500 &  & 31.26\tiny±0.67 & 14.98\tiny±0.37 &  & 38.45\tiny±0.47 & 36.72\tiny±0.70 \\
1000 &  & 36.49\tiny±0.58 & 22.44\tiny±0.57 &  & 41.08\tiny±059 & 40.27\tiny±0.69 \\\midrule
200 & \multirow{3}{*}{R-MNIST} & 92.02\tiny±0.83 & 91.74\tiny±0.83 & \multirow{3}{*}{GCIL-L} & 32.24\tiny±2.14 & 27.84\tiny±1.17 \\
500 &  & 93.37\tiny±0.45 & 93.18\tiny±0.49 &  & 33.07\tiny±2.28 & 29.60\tiny±1.14 \\
1000 &  & 93.63\tiny±0.49 & 93.51\tiny±0.48 &  & 39.96\tiny±1.59 & 39.77\tiny±1.46 \\
\bottomrule
\end{tabular}
\end{table}

\section{Comparison with CLS-ER}
CLS-ER provides us with an effective framework for studying the effect of synaptic consolidation in a multiple memory replay framework. We built our approach on top of their Mean-ER method instead of CLS-ER (which is a much stronger baseline) because of it’s simplicity and parallels with the CLS theory in the brain. While the multiple semantic memories in CLS-ER provides generalization gain, it also introduces additional complexities and makes it more challenging to study the effect of synaptic consolidation. 

The goal of our study was not to outperform the existing approaches but rather to investigate the complementary nature of synaptic consolidation and dual memory experience replay similar to the brain. Interestingly, we found that incorporating synaptic consolidation with even one semantic memory (as in Mean-ER) enables SYNERgy to perform comparable to CLS-ER (with two semantic memories) in simpler settings and provides considerable gains in the more realistic and challenging GCIL-CIFAR-100 longtail and uniform settings. Additionally, our stability plasticity analysis shows that  SYNERgy maintains a better trade-off compared to CLS-ER. CLS-ER provides higher stability largely due to the lower frequency of updates for the stable model but it comes at a considerable cost of plasticity whereas SYNERgy does not compromise the plasticity of the model. Furthermore, SYNERgy requires less computations as it involves one additional forward pass for the memory samples instead of two in CLS-ER and for the complex settings the frequency of fisher information matrix update is quite low . 

We believe that SC can further improve the performance of CLS-ER and provides an interesting research avenue for employing SC in a multiple semantic memories setup.

\section{Characteristics Analysis}
To further study the characteristics of the model, we compare the task probabilities and calibration of the models trained on different datasets with varying memory budgets.

\subsection{Model Calibration}
\begin{figure*}[t!h]
 \centering
 \includegraphics[width=0.99\textwidth]{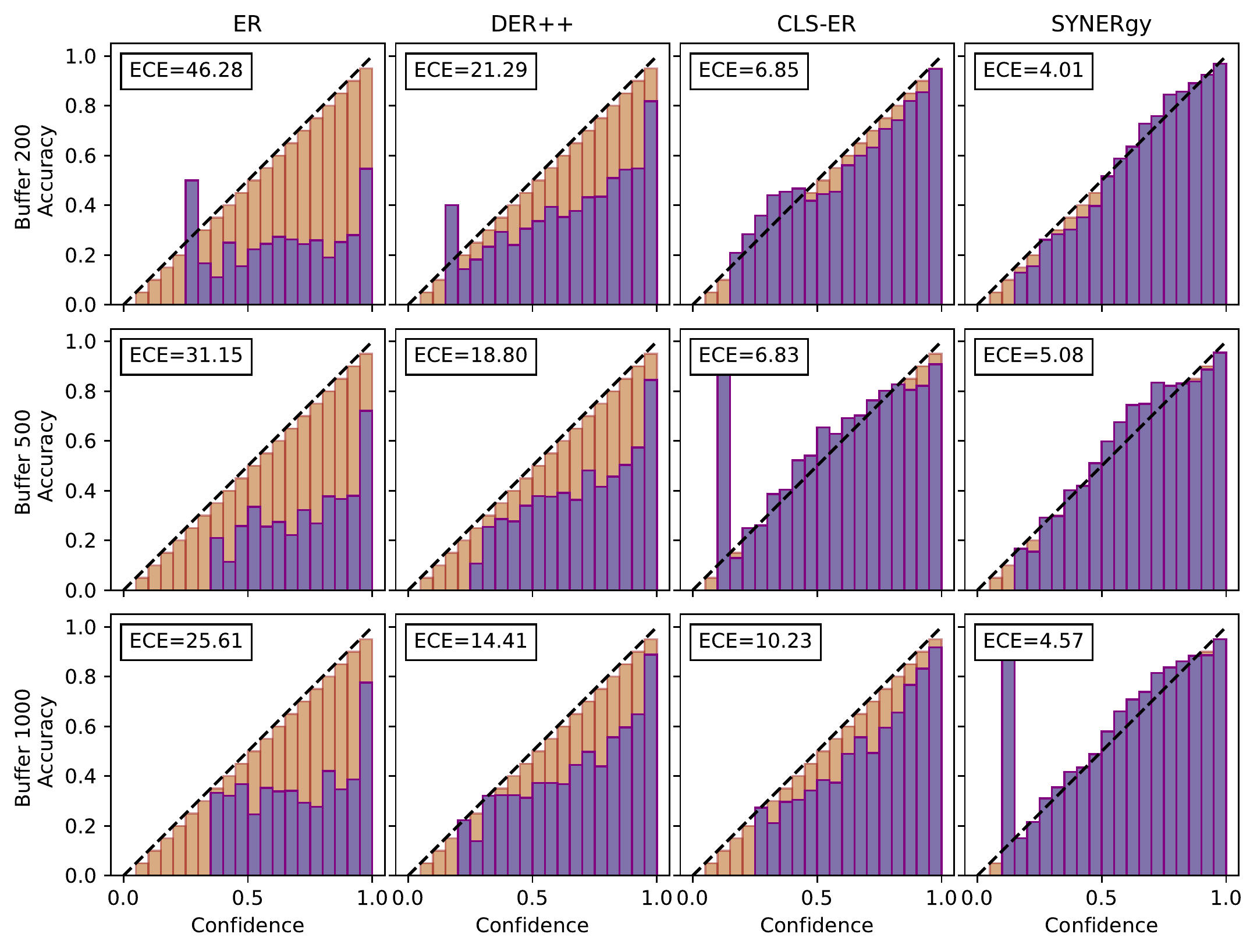}
 \caption{Reliability plots and ECE on S-CIFAR-10 with varying memory budget.}
 \label{fig:calib-cifar}
\end{figure*}

\begin{figure*}[t!h]
 \centering
 \includegraphics[width=0.99\textwidth]{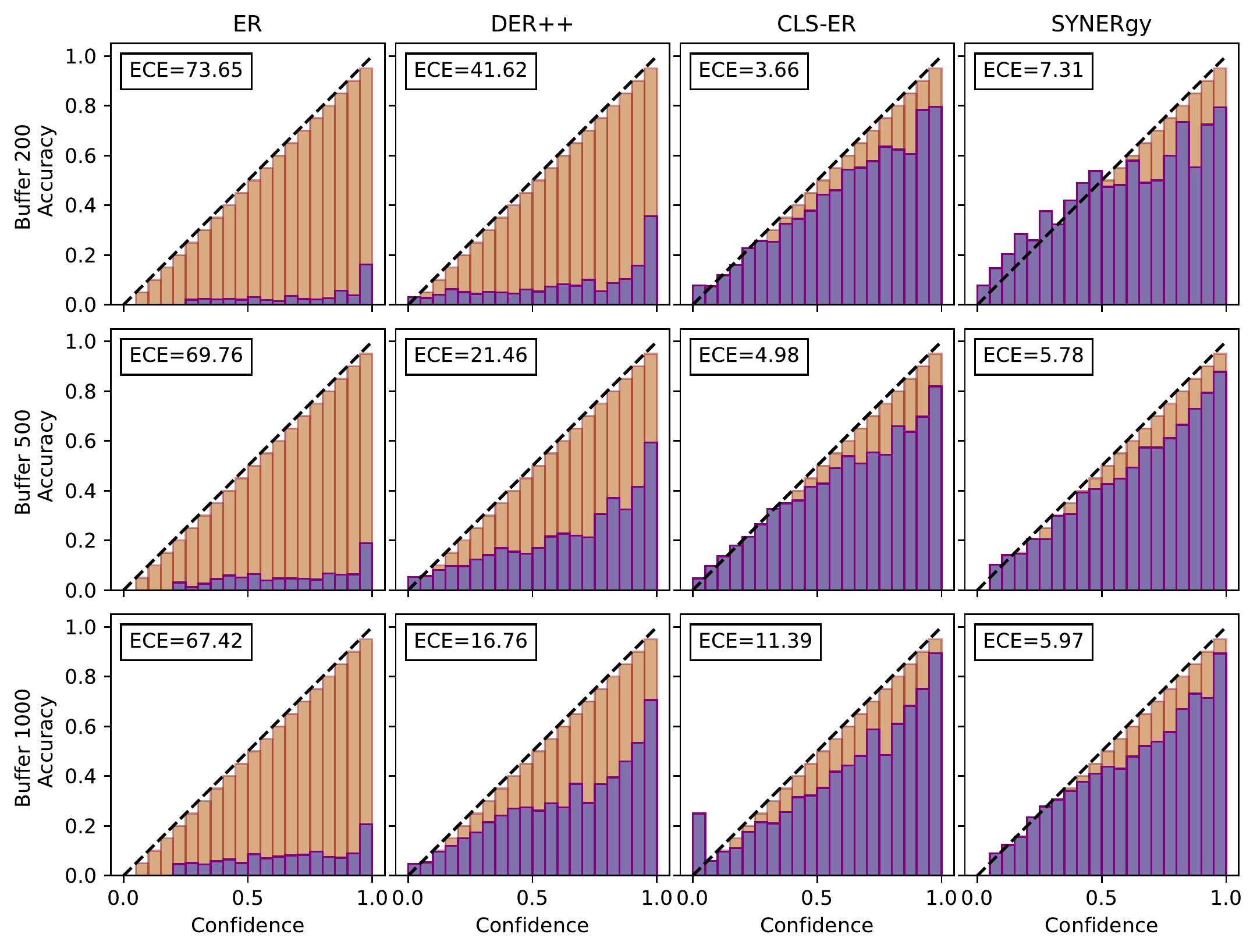}
 \caption{Reliability plots and ECE on S-TinyImageNet with varying memory budget.}
 \label{fig:calib-tinyimg}
\end{figure*}

Figures \ref{fig:calib-cifar} and \ref{fig:calib-tinyimg} provides the reliability plots and ECE of the different models trained on S-CIFAR-10 and S-TinyImageNet respectively with varying memory buffer sizes. CLS-ER and SYNERgy have significantly lower calibration error compared to ER and DER, particularly with lower buffer sizes on S-TinyImageNet. On TinyImageNet with buffer size 200 and 500, CLS-ER provides lower calibration whereas for all the other settings, SYNERgy provides lower calibration error.

\subsection{Task Probabilities}
\begin{figure*}[t!b]
 \centering
 \includegraphics[width=0.99\textwidth]{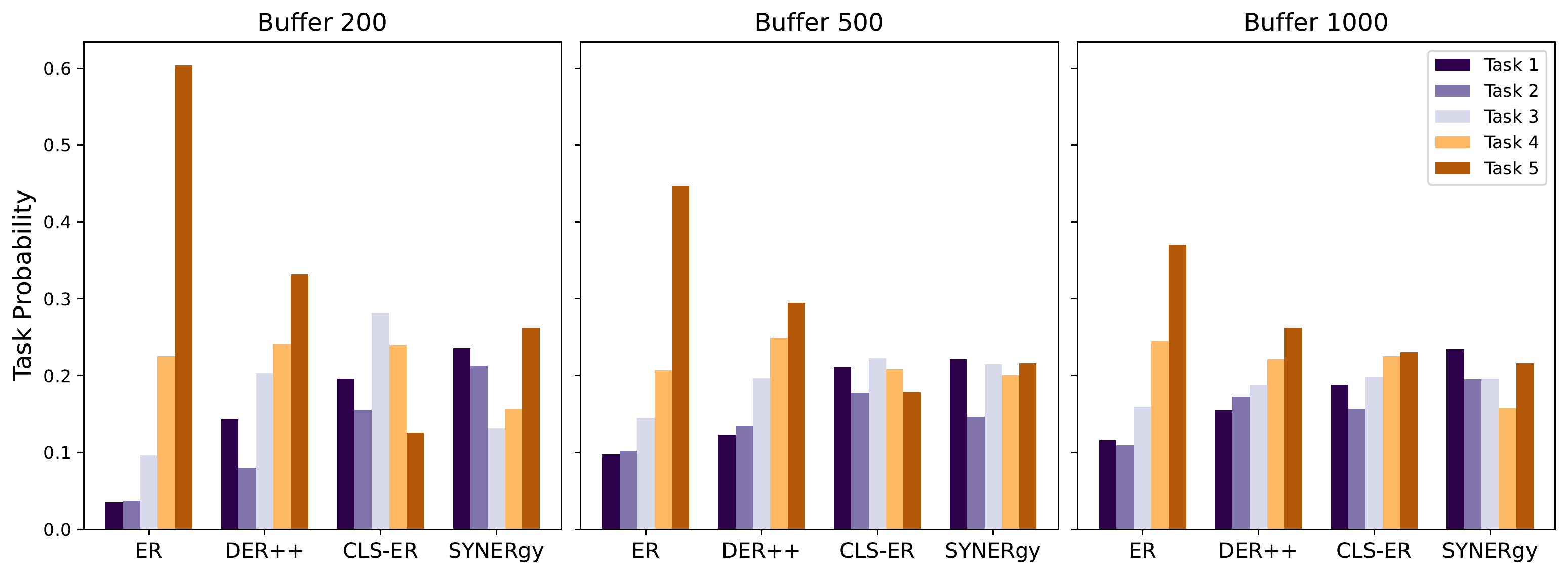}
 \caption{Task probabilities on S-CIFAR-10: SYNERgy reduces the bias towards recent task across the different memory budget.}
 \label{fig:task_prob-cifar}
\end{figure*}

\begin{figure*}[t!b]
 \centering
 \includegraphics[width=0.99\textwidth]{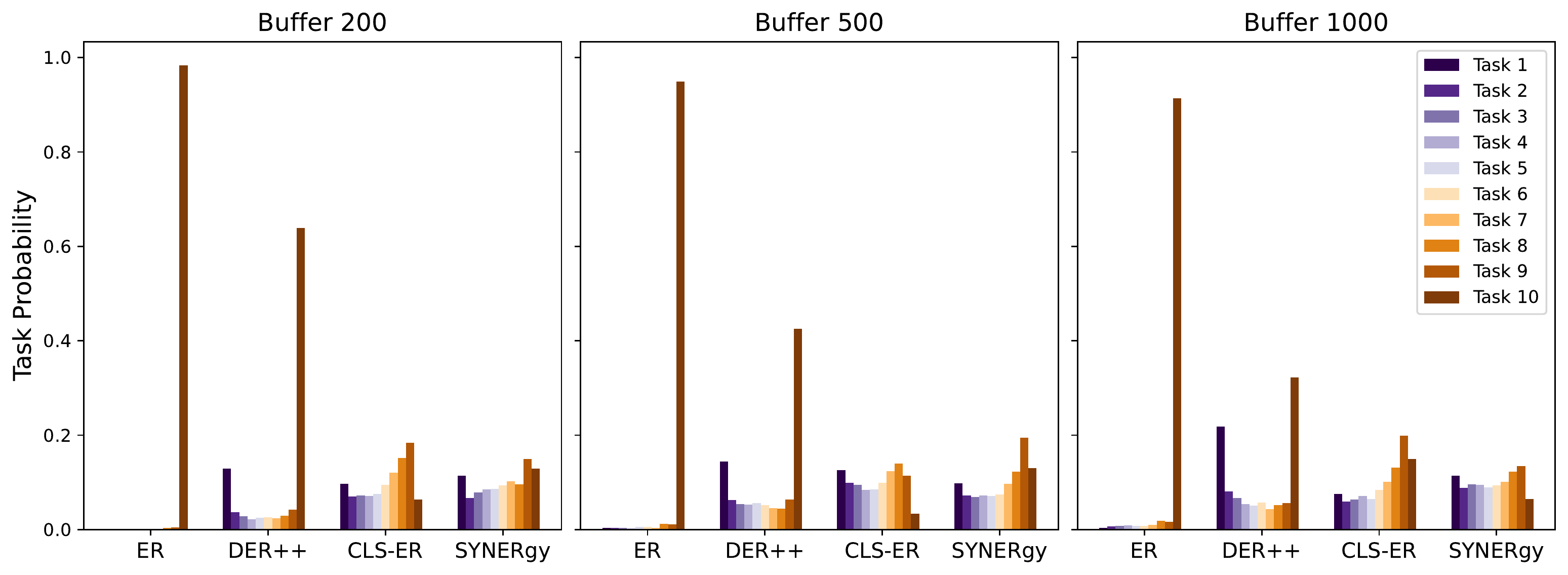}
 \caption{Task probabilities on S-TinyImageNet: SYNERgy reduces the bias towards recent task across the different memory budget.}
 \label{fig:task_prob-tinyimg}
\end{figure*}

Figure \ref{fig:task_prob-cifar} and \ref{fig:task_prob-tinyimg} provides the task probabilities of the different models trained on S-CIFAR-10 and S-TinyImageNet respectively with varying memory buffer sizes. SYNERgy reduces the bias towards the recent task and provides more uniform probability of prediction across the tasks. On S-TinyImageNet with 500 buffer size, CLS-ER has a much lower probability for the last task which is also not desirable and significantly reduces the performance of the model on the recent task.

\section{Experimental Setup and Training Details}\label{exp-setup}
To compare our method with different CL methods under common experimental settings and evaluation framework, we extended the Mammoth framework \citep{buzzega2020dark}. For each of our evaluation settings, we use the same training scheme (SGD optimizer, input stream batch size, memory buffer batch size, and the number of training epochs) and find the optimal hyperparameters for our method using a grid search over $\lambda$, $\beta$, $\alpha_{S}$, $\alpha_{F}$, $r_{S}$, and $r_{F}$ on a small validation set. The selected hyperparameters for each setting are provided in Table \ref{tab:settings}. Note that for a given dataset, the selected hyperparameters do not vary much. We use the same backbone for each dataset as in \citet{buzzega2020dark}: a fully connected network with two hidden layers, each with 100 ReLU units on  MNIST dataset variants and ResNet-18~\citep{he2015deep} for the other datasets. We apply random crop and random horizontal flip on the input stream and buffer samples for S-CIFAR10, S-TinyImageNet, and GCIL-CIFAR100. While also being interested in the online CL setting, similar to \cite{buzzega2020dark,arani2022learning}, we take into account the dataset complexity to disentangle the effect of underfitting from forgetting. We use a single epoch for MNIST-based settings whereas, for the other complex settings, we increase the number of epochs to 50 epochs. Note that CLS-ER and DER++ use 100 epochs on GCIL~\cite{arani2022learning}. For all of our experiments, we report the mean and one standard deviation of the average task accuracy of 10 differently initialized runs.

\begin{table}[ht]
\caption{The selected hyperparameters for SYNERgy in each of the experimental settings.}
\label{tab:settings}
\centering
\begin{tabular}{lcccccccccc}
\toprule
Dataset & \begin{tabular}[c]{@{}c@{}}Buffer\\ size\end{tabular} & batch\_size & \#epoch & $\eta$ & $\lambda$ & $\beta$ & $\alpha_s$ & $r_s$ & $\alpha_F$ & $r_F$ \\ \midrule
\multirow{3}{*}{S-CIFAR-10} & 200 & 32 & 50 & 0.05 & 0.2 & 5 & 0.999 & 0.2 & 0.999 & 0.01 \\
 & 500 & 32 & 50 & 0.05 & 0.2 & 5 & 0.999 & 0.4 & 0.999 & 0.007 \\
 & 1000 & 32 & 50 & 0.05 & 0.2 & 5 & 0.999 & 0.9 & 0.999 & 0.001 \\\midrule
\multirow{3}{*}{S-TinyImg} & 200 & 32 & 50 & 0.05 & 0.1 & 0.1 & 0.999 & 0.05 & 0.999 & 0.0005 \\
 & 500 & 32 & 50 & 0.05 & 0.1 & 0.1 & 0.999 & 0.05 & 0.999 & 0.0004 \\
 & 1000 & 32 & 50 & 0.05 & 0.1 & 0.1 & 0.999 & 0.05 & 0.999 & 0.0004 \\\midrule
\multirow{3}{*}{R-MNIST} & 200 & 128 & 1 & 0.2 & 1 & 1 & 0.99 & 1 & 0.99 & 0.4 \\
 & 500 & 128 & 1 & 0.2 & 1 & 1 & 0.99 & 1 & 0.99 & 0.8 \\
 & 1000 & 128 & 1 & 0.2 & 1 & 1 & 0.99 & 1 & 0.99 & 0.4 \\\midrule
\multirow{3}{*}{MNIST-360} & 200 & 16 & 1 & 0.2 & 1 & 1 & 0.99 & 0.8 & 0.99 & 0.9 \\
 & 500 & 16 & 1 & 0.2 & 1 & 1 & 0.99 & 0.8 & 0.99 & 0.8 \\
 & 1000 & 16 & 1 & 0.2 & 0.8 & 1 & 0.99 & 0.8 & 0.99 & 0.8 \\\midrule
\multirow{3}{*}{GCIL-U} & 200 & 32 & 50 & 0.05 & 0.2 & 1 & 0.999 & 0.3 & 0.999 & 0.005 \\
 & 500 & 32 & 50 & 0.05 & 0.2 & 1 & 0.999 & 0.3 & 0.999 & 0.005 \\
 & 1000 & 32 & 50 & 0.05 & 0.2 & 1 & 0.999 & 0.4 & 0.999 & 0.005 \\\midrule
\multirow{3}{*}{GCIL-L} & 200 & 32 & 50 & 0.05 & 0.2 & 1 & 0.999 & 0.2 & 0.999 & 0.005 \\
 & 500 & 32 & 50 & 0.05 & 0.2 & 1 & 0.999 & 0.2 & 0.999 & 0.005 \\
 & 1000 & 32 & 50 & 0.05 & 0.2 & 1 & 0.999 & 0.4 & 0.999 & 0.005\\
 \bottomrule
\end{tabular}
\end{table}

\subsection{GCIL-CIFAR100}
Following~\cite{arani2022learning}, we consider the uniform and longtail GCIL-CIFAR-100 setting with a sequence of 20 tasks, each with 1000 samples and a maximum of 50 classes. Because of the probabilistic sampling, different dataset seeds results in vastly different tasks. For reproducibility and fair comparison, we set the GIL-seed to 1993 for all our experiments and run with 10 differently initialized models to measure the variance of the different models on a fixed setting. We train the baselines (ER, DER++ and CLS-ER) with the hyper-parameters provided in~\cite{arani2022learning}.

\subsection{1000 Buffer Size}
We consider that the 5120 buffer size used in~\cite{arani2022learning,buzzega2020dark} for S-Cifar-10, S-TinyImageNet and Rot-MNIST to be too large for these settings to distinguish the effectiveness of the CL method. Also, there is a huge jump form the modest 500 buffer size and we believe that the performance of the model on 1000 buffer can provide more insights for the research community. We use the same hyperparameters for 1000 buffer size as reported in ~\cite{arani2022learning,buzzega2020dark} for buffer size 5120.

\section{Effect of Hyperparameters}
SYNERgy involves involves the interplay between the semantic memory and the episodic memory to enforce a consistency loss while also anchoring the working model's weights to the important parameters of the semantic memory. Given the nature of the interactions and the formulation, the different components of the method are complementary. Table \ref{tab:hyperparam} shows that our hyperparameters are complementary in nature, allowing us to keep some parameters fixed and only fine tune the other ones to facilitate hyperparamter tuning. The method is not highly sensitive to a specific set of hyperparameters for each setting as a number of settings provide similar performance.

\begin{table}[ht]
\caption{Effect of hyperparameters on SYNERgy performance (both semantic model and working model) with 500 buffer\_size, $\eta=0.05$, and $\lambda=0.2$ on GCIL-L. The values are averaged over 3 runs.}
\label{tab:hyperparam}
\centering
\begin{tabular}{c|c|c|cc||c|c|c|cc}
\toprule
$\beta$ & $r_s$ & $r_F$ & \begin{tabular}[c]{@{}l@{}}Semantic \\ memory\end{tabular} & \begin{tabular}[c]{@{}l@{}}Working \\ model\end{tabular} & $\beta$ & $r_s$ & $r_F$ & \begin{tabular}[c]{@{}l@{}}Semantic \\ memory\end{tabular} & \begin{tabular}[c]{@{}l@{}}Working \\ model\end{tabular} \\ \midrule
\multirow{9}{*}{0.5} & \multirow{3}{*}{0.2} & 0.001 & 37.25\tiny±0.50 & 32.94\tiny±0.55 & \multirow{9}{*}{1} & \multirow{3}{*}{0.2} & 0.001 & 36.66\tiny±0 & 32.69\tiny±0 \\
 &  & 0.005 & 35.09\tiny±0.21 & 32.56\tiny±0.65 &  &  & 0.005 & 33.62\tiny±2.53 & 30.08\tiny±1.62 \\
 &  & 0.010 & 36.12\tiny±0.17 & 32.06\tiny±0.82 &  &  & 0.010 & 35.99\tiny±0 & 31.87\tiny±0 \\ \cmidrule{2-5} \cmidrule{7-10}
 & \multirow{3}{*}{0.4} & 0.001 & 32.55\tiny±4.83 & 32.44\tiny±0.70 &  & \multirow{3}{*}{0.4} & 0.001 & 35.49\tiny±0 & 33.17\tiny±0 \\
 &  & 0.005 & 34.11\tiny±2.59 & 32.04\tiny±1.02 &  &  & 0.005 & 31.91\tiny±1.82 & 29.72\tiny±1.74 \\
 &  & 0.010 & 35.02\tiny±0.67 & 32.11\tiny±0.33 &  &  & 0.010 & 36.23\tiny±0 & 33.58\tiny±0 \\ \cmidrule{2-5} \cmidrule{7-10}
 & \multirow{3}{*}{0.6} & 0.001 & 33.92\tiny±0.92 & 31.95\tiny±0.66 &  & \multirow{3}{*}{0.6} & 0.001 & 33.37\tiny±1.03 & 31.21\tiny±1.01 \\
 &  & 0.005 & 33.76\tiny±1.37 & 31.66\tiny±1.05 &  &  & 0.005 & 33.67\tiny±0.76 & 31.46\tiny±0.88 \\
 &  & 0.010 & 33.02\tiny±1.33 & 31.61\tiny±0.40 &  &  & 0.010 & 33.66\tiny±0.37 & 31.97\tiny±0.39 \\
 \bottomrule
\end{tabular}
\end{table}

\end{document}